\documentclass[sigconf]{acmart}

\AtBeginDocument{%
  }

\setcopyright{acmcopyright}
\copyrightyear{2018}
\acmYear{2018}
\acmDOI{XXXXXXX.XXXXXXX}

\acmConference[Conference acronym 'XX]{Make sure to enter the correct
  conference title from your rights confirmation emai}{June 03--05,
  2018}{Woodstock, NY}
\acmPrice{15.00}
\acmISBN{978-1-4503-XXXX-X/18/06}

\usepackage{epsfig}
\usepackage{graphicx}
\usepackage{amsmath}
\usepackage{booktabs}
\usepackage{comment}
\usepackage{multirow}
\usepackage{caption}

\usepackage{subcaption}

\usepackage{color}
\usepackage{amsfonts}
\usepackage{array}
\newcolumntype{x}[1]{>{\centering\arraybackslash}p{#1pt}}

\usepackage{enumitem}
\usepackage{bm}
\usepackage{xcolor}
\usepackage{arydshln}

\usepackage{pifont}
\newcommand{\ie}{\textit{i}.\textit{e}.}
\newcommand{\eg}{\textit{e}.\textit{g}.}
\newcommand{\etc}{\textit{e}.\textit{t}.\textit{c}.}
\newcommand{\cf}{\textit{c}.\textit{f}.}

\copyrightyear{2023}
\acmYear{2023}
\setcopyright{acmlicensed}\acmConference[MMIR '23]{Proceedings of the 1st International Workshop on Deep Multimodal Learning for Information Retrieval}{November 2, 2023}{Ottawa, ON, Canada}
\acmBooktitle{Proceedings of the 1st International Workshop on Deep Multimodal Learning for Information Retrieval (MMIR '23), November 2, 2023, Ottawa, ON, Canada}
\acmPrice{15.00}
\acmDOI{10.1145/3606040.3617439}
\acmISBN{979-8-4007-0271-6/23/11}

\begin{document}

\title{Video Referring Expression Comprehension via Transformer with Content-conditioned Query}

\author{Ji Jiang}
\authornote{Both authors contributed equally to this research.}
\email{jiangji@stu.pku.edu.cn}
\affiliation{
  \institution{SECE, Peking University}
  \city{}
  \country{}
}

\author{Meng Cao}\authornotemark[1]
\email{mengcao@pku.edu.cn}
\affiliation{
  \institution{International Digital Economy Academy (IDEA)}
  \city{}
  \country{}
}

\author{Tengtao Song}
\email{songtengtao@stu.pku.edu.cn}
\affiliation{%
  \institution{SECE, Peking University}
  \city{}
  \country{}
  }

\author{Long Chen}
\email{longchen@cse.ust.hk}
\affiliation{%
  \institution{Hong Kong University of Science and Technology}
    \city{}
  \country{}
}

\author{Yi Wang}
\authornote{Corresponding author.}
\email{wygamle@gmail.com}
\affiliation{%
  \institution{Shanghai Artificial Intelligence Laboratory}
  \city{}
  \country{}
}

\author{Yuexian Zou}
\authornotemark[2]

\email{zouyx@pku.edu.cn}
\affiliation{%
  \institution{SECE, Peking University}
  \city{}
  \country{}
}

\renewcommand{\shortauthors}{Ji Jiang et al.}

\begin{abstract}
Video Referring Expression Comprehension (REC) aims to localize a target object in videos based on the queried natural language. Recent improvements in video REC have been made using Transformer-based methods with learnable queries. However, we contend that this naive query design is not ideal given the open-world nature of video REC brought by text supervision. With numerous potential semantic categories, relying on only a few slow-updated queries is insufficient to characterize them. Our solution to this problem is to create dynamic queries that are conditioned on both the input video and language to model the diverse objects referred to. Specifically, we place a fixed number of learnable bounding boxes throughout the frame and use corresponding region features to provide prior information. Also, we noticed that current query features overlook the importance of cross-modal alignment. To address this, we align specific phrases in the sentence with semantically relevant visual areas, annotating them in existing video datasets (VID-Sentence and VidSTG). By incorporating these two designs, our proposed model (called \textbf{ConFormer}) outperforms other models on widely benchmarked datasets. For example, in the testing split of VID-Sentence dataset, ConFormer achieves 8.75\% absolute improvement on Accu.@0.6 compared to the previous state-of-the-art model.

\end{abstract}



\maketitle

\section{Introduction} \label{sec:intro}
Referring Expression Comprehension (REC)~\cite{hu2017modeling,hu2016natural,yu2016modeling,yu2017joint} aims to locate the image region described by the natural language query. This task has attracted extensive attention from both academia and industry, due to its wide applications in visual question answering~\cite{antol2015vqa}, image/video analysis~\cite{anderson2018bottom,cao2022deep,mao2023improving,li2023generating}, relationship modeling~\cite{hu2017modeling,cao2023iterative}, \etc. 
Recently, with the increasing number of videos online, grounding the target object in the video domain is becoming an emerging requirement. 
Different from image REC, video REC~\cite{zhou2018weakly,vasudevan2018object,chen2019weakly,zhang2020does,feng2021siamese} is more challenging since it needs to deal with both the complex temporal and spatial information.

\begin{figure}[t]
	\centering
	\includegraphics[width=0.48\textwidth]{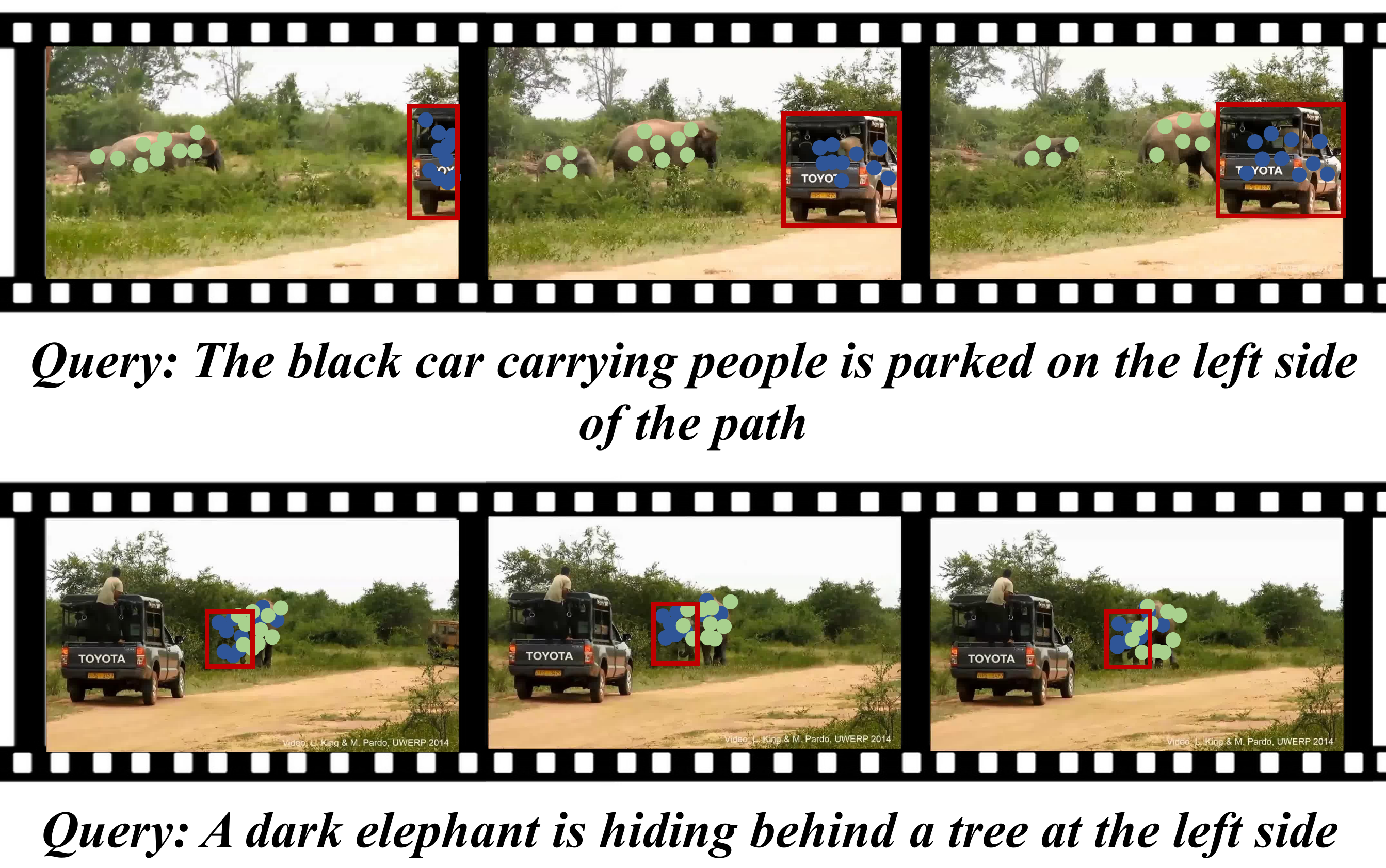}
	\caption{The distribution of our content-conditioned queries (marked in \textcolor{blue}{blue}) and traditional pure learnable queries (marked in \textcolor{green}{green}). The ground-truth annotations are marked in \textcolor{red}{red} rectangle.}
	\vspace{-0.5em}
	\label{fig:query_tube}
\end{figure}
\begin{figure}[t]
	\centering
	\includegraphics[width=0.48\textwidth]{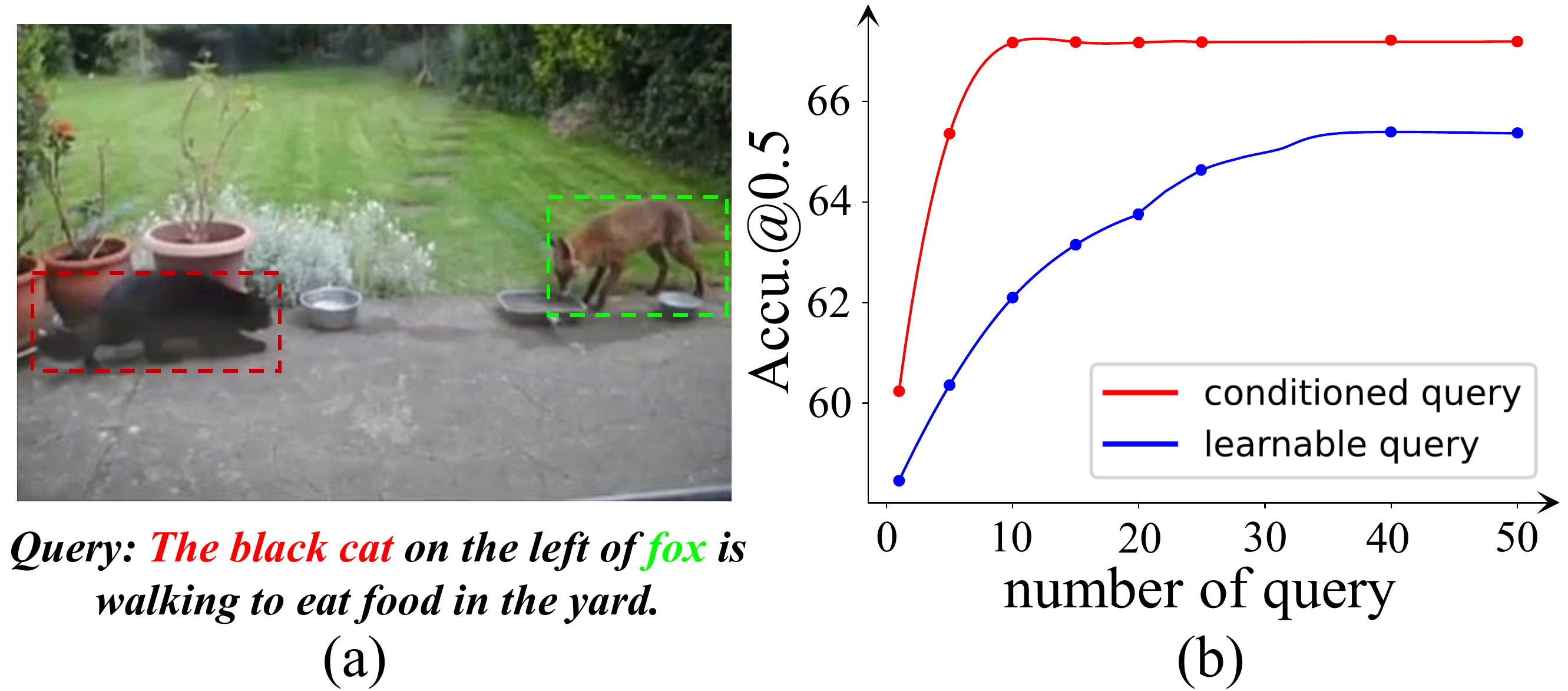}
	\caption{(a) \textbf{The motivation of fine-grained alignments.} The certain \emph{regions} (\ie, object areas) of the frame are usually more salient and highly overlapped with \emph{nouns} in the sentence containing semantic meanings. (b) \textbf{Performance \emph{v.s.} number of queries} for both our content-conditioned query and pure learnable query.}
	\label{fig:align&accu}
\end{figure}

Current mainstream methods address REC in two major directions: 1) \emph{two-stage} methods~\cite{zhang2020does,feng2021decoupled,huang2018finding,gao2017tall}: This kind of method extracts potential spatio-temporal tubes and then aligns these candidates to the sentence for finding the best matching one; 2) \emph{one-stage} methods~\cite{song2021co,sadhu2020video,zeng2020dense,chen2021end,cao2022correspondence}: They fuse visual-text features and directly predict bounding boxes densely at all spatial locations. These two kinds of methods, however, are time-consuming since they require post-processing steps (\eg, non-maximum suppression). Recently, DETR-like methods~\cite{carion2020end} have been demonstrated effective in object detection areas, which gets rid of the manually-designed rules and dataset-dependent hyper-parameters. Following this pipeline, the primary work TubeDETR~\cite{yang2022tubedetr} customizes the DETR model for video REC. Specifically, it introduces a video-text encoder and a space-time decoder, where several learnable queries are set up. Though this plain method brings noticeable performance improvements, we argue that the current query feature design for video REC is sub-optimum. Compared to object detection and segmentation, one characteristic of video REC is that it is an open-world grounding task, \ie, no pre-defined category sets are available. Therefore, an excess of queries is needed to ensure as much coverage as possible. As shown in Figure~\ref{fig:align&accu}(b), the performance is saturated when setting 40 pure learnable queries per frame.
Therefore, it is difficult to ground \emph{arbitrary} referents with pure learnable queries in our open-world setting. For example in Figure~\ref{fig:query_tube}, two scenes from the same video with different query sentences are presented. As shown, the traditional learnable query consistently focuses on the ``\texttt{elephant}" area regardless of the query sentence. In contrast, the content-conditioned query adaptively focus on the referent (``\texttt{elephant}" and ``\texttt{car}") according to both visual and language input.


Moreover, current video REC annotations only consider the correspondence between sentences and the described objects. Empirically, we contend that the nouns (\ie, subject or object) in a sentence are important to carry the overall meaning, \eg, ``\texttt{cat}" and ``\texttt{fox}" in Figure~\ref{fig:align&accu}(a) already cover the objects of interest. Therefore, the detailed alignment and differentiation between the mentioned noun objects and the corresponding visual areas (\ie, object areas or patches, \eg, the red and green dotted boxes in Figure~\ref{fig:align&accu}(a)) provide fruitful localization clues.
This asks the model to implicitly attend to keywords for matching visuals. We suppose enabling a REC model to highlight partially crucial text descriptions can make video REC more tractable. 

To this end, we propose the novel content \textbf{Con}ditioned query in Trans\textbf{former} (dubbed as \textbf{ConFormer}). We contend that the content-independent query is better adapted to the open-world grounding scenarios than the pure learnable one.
Therefore, we propose to generate query embeddings conditioned on the image content. Specifically, we set up a fixed number of bounding boxes across the frame. Then the cropped and pooled regional features are transformed into the query features of the Transformer decoder. Compared to the conventional high-dimension learnable queries, our region-based features introduce more salient prior. For example in Figure~\ref{fig:query_tube}, our content-conditioned queries adaptive focus on region-of-interest according to the input query sentence while the pure learnable one may be affected by the complex visual appearances. 

Besides, to provide the fine-grained alignment between text and videos, we contribute a semi-automatic entity generation pipeline to augment existing video REC datasets with more fine-grained text annotations, and the matched alignment losses to leverage these new supervisions. In REC annotations, we collect VID-Entity and VidSTG-Entity datasets (\cf, Figure~\ref{fig:dataset_example}) with spaCy tools and manual check, which annotate \emph{region-phrase} labels by grounding specific phrases in sentences with the bounding boxes in the video frames. To further use these detailed annotations, we also propose a fine-grained alignment loss. We firstly compute the similarity scores between each query-word pair. Then, we adopt the Hungarian algorithm~\cite{kuhn1955hungarian} to select the query matching the target bounding box. Supervised by the annotations of VID-Entity and VidSTG-Entity datasets, the InfoNCE loss is applied to map the fine-grained matched pair to be close.

We make three contributions in this paper:
\begin{itemize}[topsep=0pt, partopsep=0pt, leftmargin=13pt, parsep=0pt, itemsep=3pt]
	\item We contend that the current learnable query design can not explicitly attend to the visual and language contents, and thus not suitable for the open-world grounding scenario. To this end, we propose to generate content-conditioned queries based on the both video frame and query sentence features.
	\item Beyond the coarse-grained region-sentence one, we augment the current datasets with additional annotations and propose a fine-grained alignment loss to enhance the fine-grained \emph{region-phrase} alignment.
	\item Experimental results show that our ConFormer achieves state-of-the-art performance on both trimmed and untrimmed video REC benchmarks. 
\end{itemize}

\begin{figure*}[t]
	\centering
	\includegraphics[width=\textwidth]{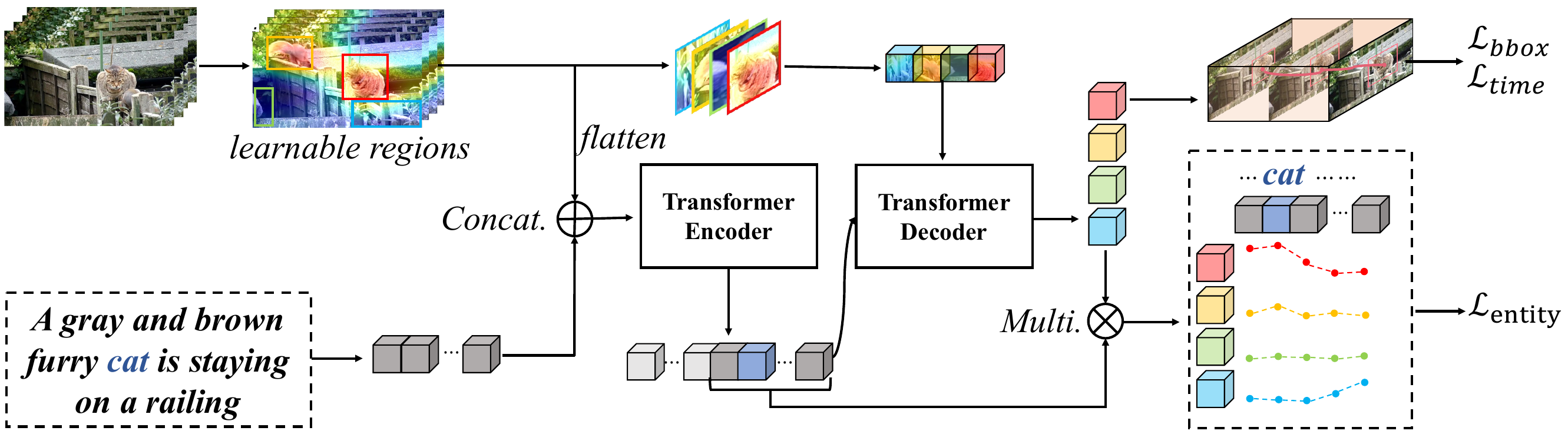}
	\caption{The schematic illustration of our ConFormer. The video and text modalities are extracted by the modality-specific backbones and fused through the Transformer encoder. A novel content-conditioned query generation module is proposed for the Transformer decoder to generate content-conditioned query features. The overall pipeline is optimized by the bi-partial matching loss and our proposed entity-aware contrastive loss.}
	\label{fig:ContFormer_pipeline}
\end{figure*}

\section{Related Work}
\noindent \textbf{Video Referring Expression Comprehension.} 
The objective of video REC is to localize the spatial-temporal tube according to the natural language query. Most of the previous works~\cite{zhang2020does,feng2021decoupled,song2021co,sadhu2020video} can be divided into two categories, \ie, two-stage and one-stage methods. However, both kinds of methods require time-consuming post-processing steps, which hinders their practical applications. 

Based on the end-to-end detection framework DETR~\cite{carion2020end}, Kamath \emph{et.al}~\cite{kamath2021mdetr} propose MDETR, an image vision-language multi-modal pre-training framework benefiting various downstream vision-language tasks. Yang \emph{et.al}~\cite{yang2022tubedetr} propose TubeDETR to conduct spatial-temporal video grounding via a space-time decoder module in a DETR-like manner. However, it still faces some problems: 1) TubeDETR processes each frame independently, which may lead to the loss of temporal information. 2) Using the pure learnable query, TubeDetr can not precisely locate the target object described by natural language sentences due to the open-world complexity. 
3) It just fuses visual and language features in a simple concatenation manner and ignores detailed vision-language alignments. In contrast, our ConFormer alleviates the above problems by introducing the content-independent query design and a fine-grained region-phrase alignment. 

\noindent \textbf{Transformer Query Design.} DETR~\cite{carion2020end} localizes objects by utilizing learnable queries to probe and filter image regions that contain the target instance. However, this learnable query mechanism is not suitable for open-world video grounding. 
To this end, current methods~\cite{liu2022dab,meng2021conditional,wang2021anchor} attempt to learn the query conditioned on the anchor points. For example, \cite{wang2021anchor} designs object queries based on anchor points to make the queries focus on anchor point areas. \cite{meng2021conditional} proposes a conditional cross-attention mechanism, which attempts to learn the conditional spatial query from decoder embedding and the reference point. \cite{liu2022dab} takes box coordinates as the queries and dynamically updates them at each layer. In our work, we employ region-of-interest features as our query design to adapt to open-world video grounding.

\noindent \textbf{Vision-Language Alignment.} 
Constructing alignment between visual and language modalities is vital in vision-language tasks. Most existing methods~\cite{radford2021learning,miech2020end,dong2019dual,cao2022locvtp,li2023g2l,zhang2022unsupervised,zhang2021cola} only build coarse-grained alignment (\eg, the image-sentence, video-sentence, region-sentence level alignment), which is not suitable for video REC. Since video REC requires localizing an instance corresponding to representative words, the alignment in video REC should be conducted in the fine-grained region-word alignment. Therefore, we contribute two new datasets with region-word annotations and propose to use these labels to regularize the model training and enhance the fine-grained alignment of the query features.

\section{Dataset Construction}
\begin{figure}[t]
	\centering
	\includegraphics[width=0.5\textwidth]{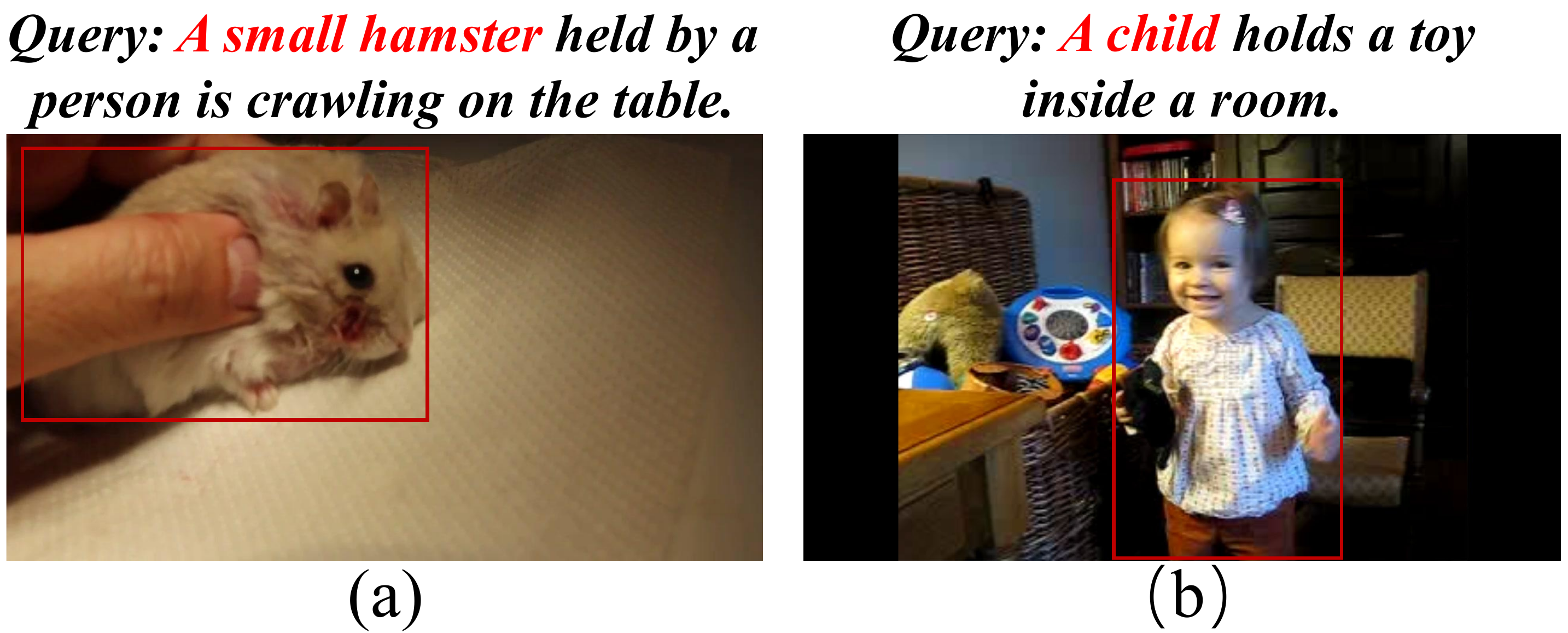}
	\caption{Visualization examples of (a) VID-Entity and (b) VidSTG-Entity datasets.}
	\label{fig:dataset_example}
\end{figure}

In this section, we give detailed illustrations of our annotated VID-Entity and VidSTG-Entity datasets. We construct them based on the widely used video REC dataset VID-sentence~\cite{chen2019weakly} and VidSTG~\cite{zhang2020does}. Specifically, VID-sentence and VidSTG contain trimmed and untrimmed videos, respectively. Beyond the existing bounding-box annotations, we explicitly annotate the words corresponding to the region-of-interest. Several examples of our datasets are illustrated in Figure~\ref{fig:dataset_example}. 

\noindent\textbf{Annotation.} To construct fine-grained word-level labels, we utilize Spacy~\cite{honnibal2015improved}, a classic natural language processing model for parsing dependency structure, to indicate the extra positional information of words corresponding to the target. Then, for quality control, we manually check and correct all the annotations. 

\noindent\textbf{Statistics.}
There are 6,582 and 80,684 region-phrase pairs in the training split of VID-Entity and VidSTG-Entity datasets, respectively. Note that the additional region-phrase annotations are only used for training. The validation and testing sets of VID-Entity and VidSTG-Entity remain the same as the original ones, which contain 536/536 spatio-temporal tubes with sentences and 8956/10302 video-sentence pairs.


\section{Method}

The schematic illustration of our ConFormer is shown in Figure~\ref{fig:ContFormer_pipeline}. Given the video-language pair as input, our objective is to output a spatial-temporal tube corresponding to the natural language query. Video and language features are extracted by the specific encoders and then are fused by the Transformer encoder (Sec~\ref{sec:3.2}). 
To adapt open-world setting, we propose a Content-conditioned Query Generation module in the decoder to generate more suitable query features instead of simply setting them as learnable. Then the Transformer decoder translates the cross-modal feature of each frame into the predicted results (Sec~\ref{sec:3.4}). The overall architecture is optimized by our proposed Entity-aware Contrastive Loss (Sec~\ref{sec:3.5}) to build the fine-grained visual-text alignment.

\subsection{Transformer Encoder}\label{sec:3.2}
The video and text features are extracted by the modality-aware backbone and projected to a shared embedding space, resulting in video features $\boldsymbol{V} \in \mathbb{R}^{T \times C \times H \times W}$ and text features $Y\in \mathbb{R}^{L\times C}$. $T$ is input frame number and $L$ is the word length. $C$ denotes the feature dimension. To ease the Transformer encoder input, we flatten $\boldsymbol{V}$ to generate  $\boldsymbol{U}  \in \mathbb{R}^{F \times C}$, where $F = T \times H \times W$. Then the encoder takes the concatenation of $\boldsymbol{U}$ and $\boldsymbol{Y}$ as input, and generates the cross-modal feature $\boldsymbol{H} \in \mathbb{R}^{ (F + L) \times C}$.
\begin{equation}
	\boldsymbol{H} = \text{MHSA}(\operatorname{Concat}(\boldsymbol{U}, \boldsymbol{Y})),
\end{equation}
\noindent where $\text{MHSA}(\cdot)$ denotes the multi-head self attention. $\operatorname{Concat}(\cdot,\cdot)$ is the concatenation operation.

\subsection{Transformer Decoder} \label{sec:3.4}

\noindent \textbf{Content-conditioned Query Generation.}
We propose a novel Content-conditioned Query Generation module to generate language-guided spatio-temporal query features to adaptively focus on the referent according to visual and language features. In Figure~\ref{fig:ContFormer_pipeline}, we firstly generate a fix number of learnable regions $\boldsymbol{R} = \left\{\boldsymbol{r}_{0}, \boldsymbol{r}_{1}, ..., \boldsymbol{r}_{N} \right\}$ for each frame and obtain the regional features $\boldsymbol{Q} = \left\{\boldsymbol{q}_{0}, \boldsymbol{q}_{1}, ..., \boldsymbol{q}_{N} \right\}$ by RoI alignment:
\begin{equation}
	\boldsymbol{q}_{i} = \operatorname{Align}(\boldsymbol{U}, r_i),
\end{equation}
\noindent where $\operatorname{Align}(\cdot, \cdot)$ is the RoI alignment operation. Note that we omit the frame index since all the frames share the same steps.

\noindent \textbf{Decoder Translation.} Given the visual-language feature $\boldsymbol{H}$ from the encoder output, we directly use the generated regional feature $\boldsymbol{Q}$ as the content-conditioned query feature. We employ the standard Transformer decoder to generate the final output $\boldsymbol{P} \in \mathbb{R}^{N \times C}$. 
\begin{equation}
	\boldsymbol{P} = \text{MHCA}(\boldsymbol{H}, \boldsymbol{Q}),
\end{equation}
\noindent where $\text{MHCA}(\cdot,\cdot)$ is the multi-head cross attention in vanilla Transformer decoder. 

To predict the temporal boundary and the bounding box sequence\footnote{Trimmed videos only require bounding box predictions.}, two multi-layer perceptrons (MLP) are applied to generate bounding box predictions $\{\boldsymbol{b}_i\}_{i=1}^{N}$ and temporal predictions $\{\boldsymbol{p}_i\}_{i=1}^{N}$, where $\boldsymbol{b}_i \in \mathbb{R}^{4}$ is the bounding box coordinates and $\boldsymbol{p}_i \in \mathbb{R}^{2}$ represents the frame-wise possibility of the start and end frames.

\subsection{Training Objectives}\label{sec:3.5}
The optimization target of our model consists of a bi-partial matching loss and our proposed entity-aware contrastive one as:
\begin{equation}
	\mathcal{L} = \mathcal{L}_{\text{match}} + \lambda_{\text{entity}} \mathcal{L}_{\text{entity}},
\end{equation}
\noindent where $\lambda_{\text{entity}}$ is the balancing factor.

\noindent\textbf{Bi-partial Matching Loss.} We split the cross-modal feature $\boldsymbol{H} \in \mathbb{R}^{ (F + L) \times C}$ to the attended visual feature $\boldsymbol{H}^{V} \in \mathbb{R}^{ F \times C}$ and text feature $\boldsymbol{H}^{Y} \in \mathbb{R}^{L \times C}$. To find the matching query in our \emph{many-to-one} setting~\cite{cao2021pursuit}, we employ the bi-partial matching strategy. We firstly select the query item with the minimum costs as follows. 
\begin{equation}
	i^* = \underset{i \in [1, N]}{\arg \min}[-\log \boldsymbol{p}_{i} + \mathcal{L}_{\mathrm{box}} + \mathcal{L}_{\mathrm{time}}],
	\label{eq:match1}
\end{equation}
\noindent where $\mathcal{L}_{\mathrm{box}}$ and $\mathcal{L}_{\mathrm{time}}$ are the bounding box regression loss and temporal boundary loss, respectively. 

$\mathcal{L}_{\mathrm{box}}$ is implemented as follows. 
\begin{equation}
	\mathcal{L}_{\mathrm{box}} =\lambda_{\text{giou}}\mathcal{L}_{\text{giou}} +\lambda_{L1}\lvert\lvert \boldsymbol{b}-\hat{\boldsymbol{b}}\rvert\rvert_1,
\end{equation}
\noindent where $\mathcal{L}_{\text{giou}}$ is the scale-invariant generalized intersection over union~\cite{rezatofighi2019generalized}.

$\mathcal{L}_{\mathrm{time}}$ is the temporal boundary loss, which is only used in training untrimmed video datasets (\eg, VidSTG-Entity). We use the Kullback-Leibler divergence values to evaluate the temporal predictions. $\hat{\boldsymbol{b}}$ is the ground truth temporal annotations.
\begin{equation}
\mathcal{L}_{\text{time}}=\lambda_{\text{KL}}\mathcal{L}_{\text{KL}}(\boldsymbol{b}, \hat{\boldsymbol{b}}).
\end{equation}

The overall matching loss is computed as follows.
\begin{equation}
	\mathcal{L}_{\text{match}} = -\log \boldsymbol{p}_{i^*} + \mathcal{L}_{\mathrm{box}}(\boldsymbol{b}_{i^*}) + \mathcal{L}_{\mathrm{times}}(\boldsymbol{b}_{i^*}).
	\label{eq:match}
\end{equation}

\noindent\textbf{Entity-aware Contrastive Loss.} 

Benefiting from the fine-grained region-phrase annotations in our proposed VID-Entity and VidSTG-Entity datasets, we build the detailed entity-aware contrastive loss to pull the word and the corresponding regional features to be close.
\begin{equation}
	\mathcal{L}_{\text{entity}} = - \log \left(\frac{\exp  (\boldsymbol{H}^V_{i^*})^\text{T} \boldsymbol{H}^Y_{+} / \tau}{\sum^{L}_{k=1} \exp (\boldsymbol{H}^V_{i^*})^\text{T} \boldsymbol{H}^Y_{k}/\tau}\right),
\end{equation}
\noindent where $\boldsymbol{H}^Y_{+}$ is the matched positive sample for $\boldsymbol{H}^V_{i^*}$, $\tau$  is a temperature parameter.

\section{Experiments}
We firstly introduce the experimental settings in Sec.~\ref{sec:5.1}. Then, we compare our ConFormer with the current state-of-the-art methods in Sec.~\ref{sec:5.2} on two video datasets and an image one. In Sec~\ref{sec:5.3}, we further verify the effectiveness of each proposed module. Finally, several visualization results are presented along with analysis in Sec~\ref{sec:5.4}.

\begin{table*}[t]

\caption{Comparison (\%) with state-of-the-art methods on VidSTG-Entity dataset. (VG: Visual Genome~\cite{krishna2017visual}, CC: Conceptual Captions~\cite{sharma2018conceptual}, IN: ImageNet~\cite{deng2009imagenet}, F: Flickr30k~\cite{plummer2015flickr30k}, C:COCO~\cite{lin2014microsoft}). PT data denotes pretraining data.}
\renewcommand\arraystretch{1.1}
\setlength{\tabcolsep}{5pt}

\begin{center}
\resizebox{0.9\linewidth}{!}{
\begin{tabular}{lcccccccccccccccccccccc}
\toprule
\multirow{2}*{\textbf{Method}} &\multirow{2}*{\textbf{PT Data}}&& \multicolumn{4}{c}{\textbf{Declarative Sentences}}&& \multicolumn{4}{c}{\textbf{Interrogative  Sentences}} \\
\cmidrule{4-7}
\cmidrule{9-12}
&&& m\_tIoU& m\_vIoU & vIoU@0.3 & vIoU@0.5&& m\_tIoU& m\_vIoU & vIoU@0.3 & vIoU@0.5 \\
\midrule
STGRN~\cite{zhang2020learning} & VG &&\textbf {48.5} &19.8 & 25.8 & 14.6 && \textbf{47.0} & 18.3 & 21.1 &12.8 \\
STGVT~\cite{su2019vl} & VG+CC && - &  21.6  & 29.8 & 18.9 && - & - & - & - \\
STVGBert~\cite{su2021stvgbert} & IN+VG+CC && - &  24.0  &  30.9 &  18.4  && - &  22.5 & 26.0 & 16.0\\
TubeDETR~\cite{yang2022tubedetr} & IN && 43.1 &  28.0  &  39.9 &  26.6  && 42.3 &  25.1 & 35.7 & 22.4\\
ConFormer (Ours) & IN && 45.0 &  29.9  &  40.1 &  27.8  && 43.6 & 26.5 & 36.0 & 23.5 \\
Tubedetr~\cite{yang2022tubedetr}&IN+VG+F+C& &48.1&30.4 &42.5  &28.2 &&46.9 & 25.7&35.7&23.2 \\

ConFormer (Ours) & IN+VG+F+C && 48.3 &  \textbf{31.8}  &  \textbf{44.3} & \textbf{ 30.5}  && \textbf{47.0} & \textbf{27.7} &\textbf{ 37.6} & \textbf{24.8} \\

\bottomrule
\end{tabular}
}
\end{center}

\label{table:vidSTG-entity}
\end{table*}

\begin{table}[h]
\caption{Comparison (\%) with state-of-the-art methods on VID-Entity dataset.}
\renewcommand\arraystretch{1.1}

\begin{center}
\resizebox{0.8\linewidth}{!}{
\begin{tabular}{ccccccccccccccc}
\toprule
\multirow{2}*{\textbf{Method}} & \multicolumn{3}{c}{\textbf{Accu.@}} \\
\cmidrule{2-4}
& 0.4 & 0.5 & 0.6  \\
\midrule
Yang \emph{et al.} (\emph{w/} BERT)~\cite{yang2019fast} & - & 52.39 & -  \\
Yang \emph{et al.} (\emph{w/} LSTM)~\cite{yang2019fast} & - & 54.78 & -  \\
DVSA+Avg~\cite{karpathy2015deep} & 36.2 & 29.7 & 23.5 \\
DVSA+NetVLAD~\cite{karpathy2015deep} & 31.2 & 24.8 & 18.5  \\
DVSA+LSTM~\cite{karpathy2015deep} & 38.2  & 31.2 & 23.5  \\
GroundeR+Avg~\cite{rohrbach2016grounding} & 36.7 & 31.9 & 25.0 \\
GroundeR+NetVLAD~\cite{rohrbach2016grounding} & 26.1 & 22.2 & 15.1  \\  
GroundeR+LSTM~\cite{rohrbach2016grounding} & 36.8 & 31.2 & 27.1  \\
\cdashline{1-4}[2pt/2pt]
First-frame tracking~\cite{li2019siamrpn++} & - & 36.97 & -  \\
Middle-frame tracking~\cite{li2019siamrpn++} & - & 44.00 & - \\
Last-frame tracking~\cite{li2019siamrpn++} & - & 36.26 & -  \\
Random-frame tracking~\cite{li2019siamrpn++} & - & 40.20 & -\\
\cdashline{1-4}[2pt/2pt]
WSSTG~\cite{chen2019weakly} & 44.60 & 38.20 & 28.90 \\
Co-grounding~\cite{song2021co} & 63.35  & 60.25 & 53.89  \\
\cdashline{1-6}[2pt/2pt]
\textbf{ConFormer (Ours)} & \textbf{70.40} & \textbf{67.27} & \textbf{62.64} \\
\bottomrule
\end{tabular}
}
\end{center}

\label{table:VID-entity}
\end{table}


\subsection{Experimental Settings}\label{sec:5.1}

\noindent\textbf{Evaluation Metric.} For the trimmed video dataset, \ie, VID-Entity, we employ the bounding box localization accuracy Accu.@$\eta$, where a predicted result is considered correct if the IoU between the predicted region and ground-truth region is greater than the threshold $\eta$. For the untrimmed video dataset, \ie, VidSTG -Entity, we follow~\cite{zhang2020does} to adopt m\_tIoU, m\_vIoU, and vIoU@$\theta$ as our evaluation criteria. m\_tIoU is the average temporal IoU between the predicted timestamps and the ground-truth annotations. Following \cite{zhang2020does}, vIoU is defined as $\frac{1}{|S_U|}\sum_{t\in S_I}r_t$, where $r_t$ is the IoU between the predicted bounding box and ground-truth bounding box at the $t^{th}$ frame. $S_I$ is the intersections of the predicted tubes and ground-truth tubes and $S_U$ is the union of them. m\_vIoU is the average of vIoU and vIoU@$\theta$ is the average ratio of samples with vIoU greater than threshold $\theta$. In this paper, we set $\eta$ to 0.4, 0.5, 0.6 and $\theta$ to 0.3, 0.5. 

\noindent\textbf{Implementation details.} For both datasets, we decoded the video by setting \emph{fps} to 5. For the VID-Entity dataset, we set the input frame number $T$ to 20 and the longer edge length to $672$. Since the untrimmed video dataset VidSTG-Entity requires temporal localization, we set $T$ to 200 for the ease of localization. The longer edge length in VidSTG-Entity is set to $256$. We used ResNet101~\cite{he2016deep} pretrained on ImageNet~\cite{deng2009large} as our visual backbone and RoBERTa~\cite{liu2019roberta} pretrained from HuggingFace~\cite{wolf2019huggingface} as the text encoder. We used the AdamW~\cite{loshchilov2017decoupled} optimizer with the initial learning rate setting to $10^{-4}$. The training process lasted for 10 epochs on both datasets. We set $\lambda_{giou}$, $\lambda_{L1}$, $\lambda_{KL}$, $\lambda_{entity}$ and  $\tau$  to 2, 5, 5, 1, and 0.07, respectively.

\begin{table}[t] 
\caption{Ablation studies on trimmed video dataset VID-Entity and untrimmed video dataset VidSTG-Entity, respectively. (VTCQ: Video-Text Conditioned Query Generation, ECL: Entity-aware Constrastive Loss)}

\renewcommand\arraystretch{1.1}
\setlength{\tabcolsep}{2pt}

\begin{center}
\resizebox{0.85\linewidth}{!}{
\begin{tabular}{lcccccccccccccc}
\toprule
\multirow{2}*{\textbf{Mode}} &\multirow{2}*{\textbf{VTCQ}}&\multirow{2}*{\textbf{ECL}} &\multicolumn{2}{c}{\textbf{VID-Entity}} && \multicolumn{2}{c}{\textbf{VidSTG-Entity}}\\
\cmidrule{4-5}
\cmidrule{7-8}

&&&Accu.@0.5& m\_IoU&&m\_tIoU&m\_vIoU  \\
\midrule
\#1&\ding{51} & \ding{51} & \textbf{67.17}&\textbf{0.600}&&\textbf{44.21}&\textbf{28.08}\\
\#2&\ding{55} & \ding{51} & 65.60&0.596&&43.01&27.38\\
\#3&\ding{51} & \ding{55} & 62.10&0.551&&41.72&25.56\\
\#4&\ding{55} & \ding{55} & 61.17&0.543&&41.05&24.77\\

\bottomrule
\end{tabular}
}
\end{center}

\label{table:ablation}
\end{table}

\subsection{Comparisons with State-of-the-Arts}\label{sec:5.2}
\begin{figure}[t]
	\centering
	\includegraphics[width=0.48\textwidth]{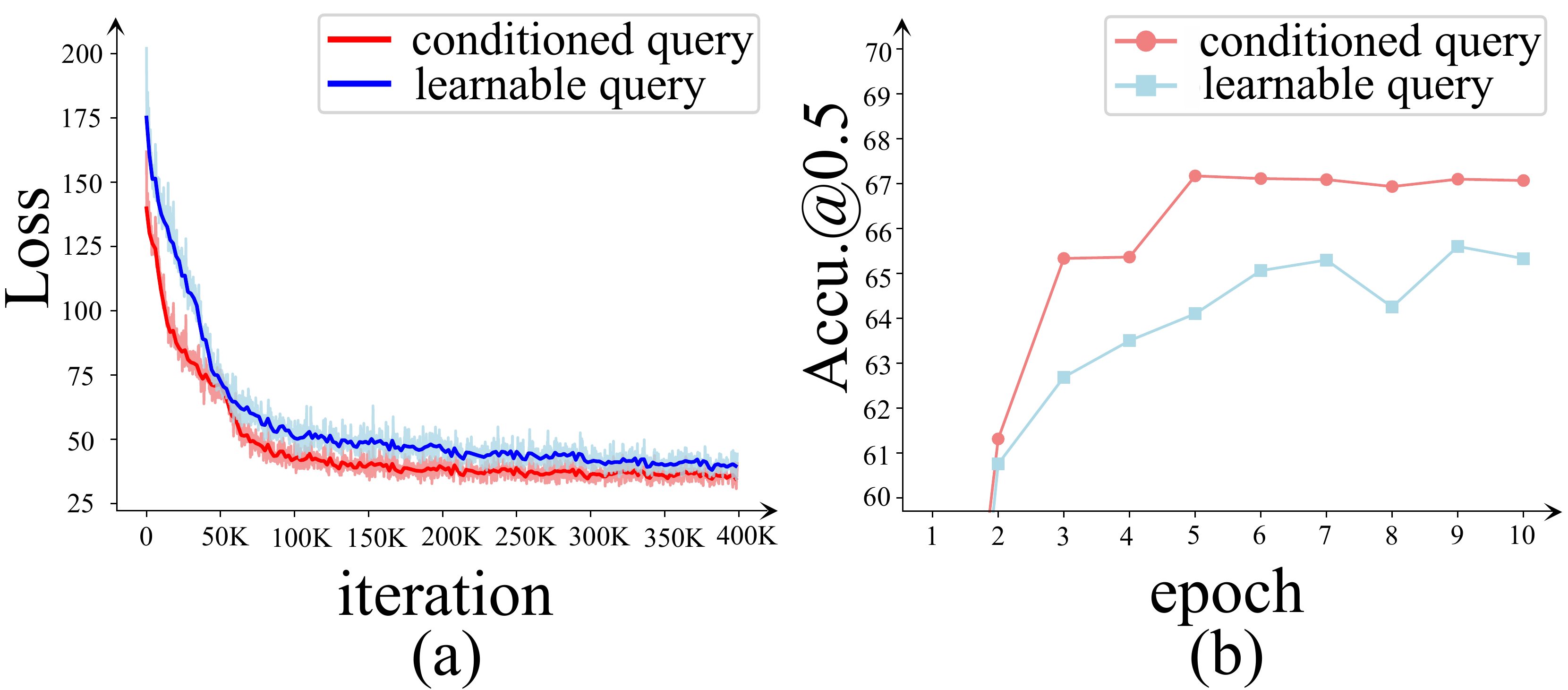}
	\caption{(a) Comparison of loss curve of content conditioned query (marked in \textcolor{red}{red}) and pure learnable query (marked in \textcolor{blue}{blue}). (b) Comparison of Accu.@0.5 between video-text conditioned query (marked in \textcolor{red}{red}) and pure learnable query (marked in \textcolor{blue}{blue}).}
	\label{fig:loss_iteration_accu}
\end{figure}

\noindent\textbf{Results on VID-Entity dataset.} 
We compare our ConFormer with state-of-the-art methods on VID-Sentence dataset in Table.~\ref{table:VID-entity}. We classify the compared methods into three categories: \textbf{1) Grounding frame-wisely}. We transfer state-of-the-art image REC methods (\ie, Yang \emph{et.al}~\cite{yang2019fast}, DVSA~\cite{karpathy2015deep}, and GroundeR~\cite{rohrbach2016grounding}) to the video scenario by adding the temporal interaction module (\ie, Avg, NetVLAD~\cite{arandjelovic2016netvlad} and LSTM). 2) \textbf{Tracking frame-wisely}: We attempt to utilize the state-of-the-art tracker~\cite{li2019siamrpn++} to solve video REC. Specifically, we acquire the tracking template on the first/middle/last/random frame by one-stage LSTM~\cite{yang2019fast}, then tracker~\cite{li2019siamrpn++} is applied to track the template according to frames.  3) \textbf{Other video REC methods}: We make comparisons with state-of-the-art video REC methods including  WSSTG~\cite{chen2019weakly}, and Co-grounding~\cite{song2021co}. As shown in Table~\ref{table:VID-entity}, our ConFormer achieves huge improvement boosts. For example, we achieve 7.05\%,7.02\%, and 8.75\% absolute improvement on Accu.@0.4, Accu.@0.5 and Accu.@0.6, respectively. The comparison results demonstrate that our ConFormer has made substantive progress in video REC. 

\noindent\textbf{Results on VidSTG-Entity dataset.} 
In Table~\ref{table:vidSTG-entity}, we also conduct experiments on the untrimmed video dataset VidSTG-Entity to further explore the effectiveness and generalization of ConFormer. For temporal tube prediction in untrimmed videos, we only utilize a simple temporal boundary loss to realize it, instead of elaborately designing a time-aligned cross-attention module in Tubeder~\cite{yang2022tubedetr} or constructing a well-designed 2D temporal feature map in STGRN~\cite{zhang2020learning}. Nonetheless, our ConFormer achieves a greater performance on vIoU@0.3, vIoU@0.5 and m\_vIoU than other state-of-the-art video methods (\ie, STGRN~\cite{zhang2020learning}, STGVT~\cite{su2019vl}, STVGBert~\cite{sharma2018conceptual}, and TubeDETR~\cite{yang2022tubedetr}).

\begin{table}[t]
\caption{Comparison (\%) with state-of-the-art methods on RefCOCO dataset.}
\renewcommand\arraystretch{1.1}

\begin{center}
\resizebox{0.7\linewidth}{!}{
\begin{tabular}{ccccccccccccccc}
\toprule
\multirow{2}*{\textbf{Method}} &\multicolumn{3}{c}{\textbf{RefCOCO}} \\
\cmidrule{2-4}
& val & testA & testB \\
\midrule
VC~\cite{zhang2018grounding} & - & 73.33 & 67.44  \\
ParalAttn~\cite{zhuang2018parallel}& - & 75.31 & 65.52 \\
MAttNet~\cite{yu2018mattnet} & 76.65 & 81.14 & 69.99  \\
DGA~\cite{yang2019dynamic} & - & 78.42  & 65.53  \\
FAOA~\cite{yang2019fast} & 72.54 & 74.35 & 68.50  \\
NMTree~\cite{liu2019learning}& 76.41 & 81.21  & 70.09  \\
ReSC-Large~\cite{yang2020improving} & 77.63 & 80.45 & 72.30  \\
UNITER\_L~\cite{chen2019uniter} & 81.41 & 87.04 & 74.17  \\
VILLA\_L~\cite{gan2020large} & 82.39 & 87.48 & 74.84 \\
TransVG~\cite{deng2021transvg} & 81.02 & 82.72 & 78.35  \\
RefTR~\cite{li2021referring} & 85.65 & 88.73 & 81.16 \\
VGTR~\cite{du2022visual} & 79.20 & 82.32 & 73.78  \\
SeqTR~\cite{zhu2022seqtr} & 87.00 & 90.15 & \textbf{83.59} \\
\cdashline{1-6}[2pt/2pt]
\textbf{ConFormer} & \textbf{87.72} & \textbf{90.35} & 83.36 \\
\bottomrule
\end{tabular}
}
\end{center}

\label{table:refCOCO}
\end{table}


\noindent\textbf{Results on RefCOCO dataset.} 
Besides experiments on videos, we also validate the effectiveness of our ConFormer in the image domain. For the image, we treat it as a video with only one frame and input it into our framework. As shown in Table~\ref{table:refCOCO}, we compare our model with state-of-the-art image Referring expression comprehension (REC) methods. Although our method is not specifically designed for images, our method still outperforms previous methods due to the effectiveness of VTCQ.

\subsection{Ablation studies}\label{sec:5.3}
\begin{table*}[h]
\caption{(a) Comparison of performance with or without Entity-aware Contrastive Loss (ECL).  (b) Comparison of performance with various $T$ and spatial resolution\footref{note:resolution} on the VID-Entity testing set. (c) Comparison of performance with various maximum of $T$ and spatial resolution on the VidSTG-Entity testing set.}
\centering

\renewcommand\arraystretch{1.1}
\setlength{\tabcolsep}{5pt}
    \begin{subtable}[h]{0.32\textwidth}
        \centering
        \begin{tabular}{ccc}
        \toprule
        \textbf{Method}  & \textbf{Accu.@0.5} \\
        \midrule
        One-stage LSTM~\cite{yang2019fast} & 54.78  \\
        + ECL & 56.13 \\
        \cdashline{1-3}[2pt/2pt]
        Co-grounding~\cite{song2021co}& 60.25  \\
        + ECL &63.88\\
        \bottomrule
        \end{tabular}
        \caption{}
        \label{table:ECL}
     \end{subtable}\hfill
    \begin{subtable}[h]{0.32\textwidth}
        \centering
        \begin{tabular}{cccc}
        \toprule
        \textbf{$T$} & \textbf{Resolution} & \textbf{Accu.@0.5}& \\
        \midrule
        15  & 800 & 66.38&  \\
        20 & 672 &\textbf{67.17}&\\
        25 & 576 &65.93&\\
        35 & 480 &59.28&\\
        \bottomrule
        \end{tabular}
        \caption{}
        \label{table:T_R_A}
     \end{subtable}\hfill
    \begin{subtable}[h]{0.32\textwidth}
        \centering
        \begin{tabular}{cccc}
        \toprule
        \textbf{$T$} & \textbf{Resolution} & \textbf{m\_tIoU} & \textbf{m\_vIoU} \\
        \midrule
         100 & 288&42.16&27.31 \\
         150 & 256 &43.33&27.56\\
         200 & 224 &44.21&28.08\\
         200 & 256 &\textbf{44.35}&\textbf{28.29}\\
        \bottomrule
        \end{tabular}
        \caption{}
        \label{table:T_R_M}
     \end{subtable}\hfill

\label{table:E_T_R_M}
\end{table*}

\begin{figure}[t]
	\centering
	\includegraphics[width=0.45\textwidth]{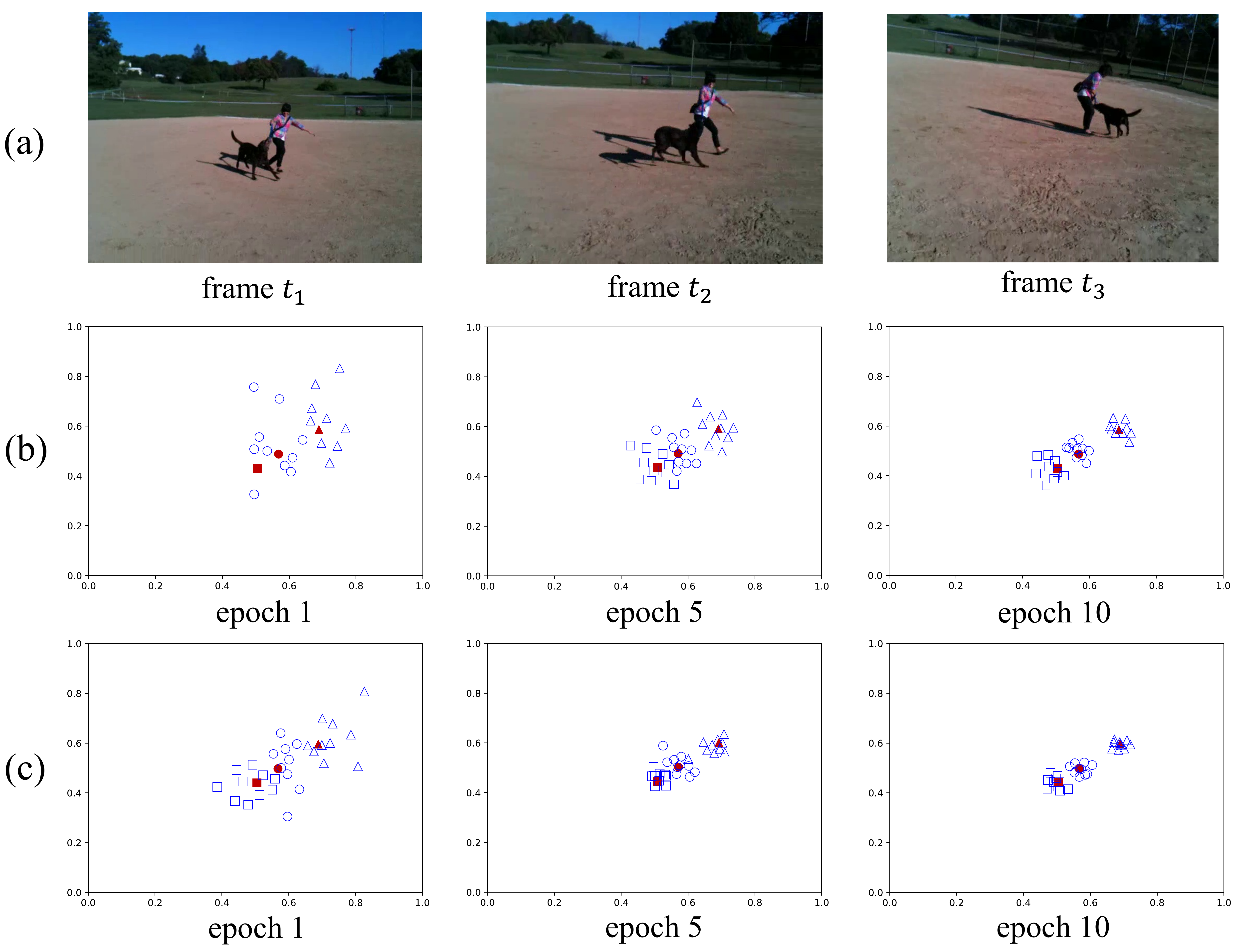}
    \caption{(a) Input image frames. The distribution of the queries generated by our model in different epochs (b) \emph{w/o} and  (c) \emph{w/} VTCQ, respectively. The localization of the target object in each frame is marked in \textcolor{red}{red}, the generated queries between predicted temporal boundaries are marked in \textcolor{blue}{blue}.}
	\label{fig:query_dis}
\end{figure}

\noindent\textbf{Content-conditioned Query Generation.} We train our model with or without VTCQ. The loss curve comparisons are shown in Figure~\ref{fig:loss_iteration_accu}(a). Note the loss curve converges faster with VTCQ. Moreover, as illustrated in Figure~\ref{fig:loss_iteration_accu}(b), the model without VTCQ achieves its best performance (with Accu.@0.5 reaching 65.60\%) at the ninth epoch. In contrast, it only requires 5 epochs to reach 65.60\% (Accu.@0.5) with VTCQ. For quantitative analysis, as shown in Table~\ref{table:ablation}, the performance with VTCQ has been improved in both VID-Entity and VidSTG-Entity. For qualitative analysis, we visualize the distribution of queries with and without VTCQ. As illustrated in Figure~\ref{fig:query_dis}, after training the first epoch, the model without VTCQ mistakes the frame $t_1$ as the background frame and generates no query for frame $t_1$. In the contrast, the model with VTCQ predicts the correct temporal boundary. In addition, the queries generated by the model with VTCQ are obviously more convergent to the target point. These experimental results demonstrate the effectiveness of our VTCQ, which can achieve better performance with fewer iterations.

\noindent\textbf{Entity-aware Contrastive Loss.} We further conduct ablation studies of Entity-aware Contrastive Loss (ECL) on VID-entity and VidSTG-entity. Results are summarized in Table~\ref{table:ablation}. On VID-Sentence, ECL brings about 4.43\% and 0.053 improvements in Accu.@0.5 and m\_IoU, respectively. In terms of m\_tIoU and m\_vIoU, it achieves 1.96\% and 2.61\% gains on VidSTG. The significant improvement in spatial localization shows that our ECL successfully guides our model to learn a better region-word alignment. 
\begin{table}[t]
\caption{Comparison performance with different approaches of integration.}
\renewcommand\arraystretch{1.1}
\begin{center}
\resizebox{0.65\linewidth}{!}{
\begin{tabular}{cccccccccccccc}
\toprule
\multirow{2}*{\textbf{Mode}} & \multicolumn{2}{c}{\textbf{VID-Entity}} \\
\cmidrule{2-3}
& Accu.@0.5 &m\_IoU \\
\midrule
Cross-Attention & 64.83 & 0.579   \\
Concat & \textbf{67.17} & \textbf{0.600}  \\
\bottomrule
\end{tabular}
}
\end{center}

\label{table:fuse}
\end{table}


\noindent\textbf{Generality of ECL.} To study the generality of our ECL, we retrain the compared methods based on our collected dataset and ECL. Specifically, we select the region feature of the target object (\ie, In one-stage methods, we first fix the receptive field and position of the object according to annotated bounding boxes. Then, we select the corresponding feature in their backbone). We choose phrase features directly related to the target, take the region-phrase pair as the positive samples, and apply our proposed entity-aware contrastive loss to construct the fine-grained alignment. As shown in Table~\ref{table:E_T_R_M}(a), the results with ECL have been all improved, which validates the effectiveness of our annotations and ECL. Besides, under the same supervision, our ConFormer still performs much better than existing methods. It shows the superiority of our architecture.

\begin{figure}[t]
	\centering
	\includegraphics[width=0.38\textwidth]{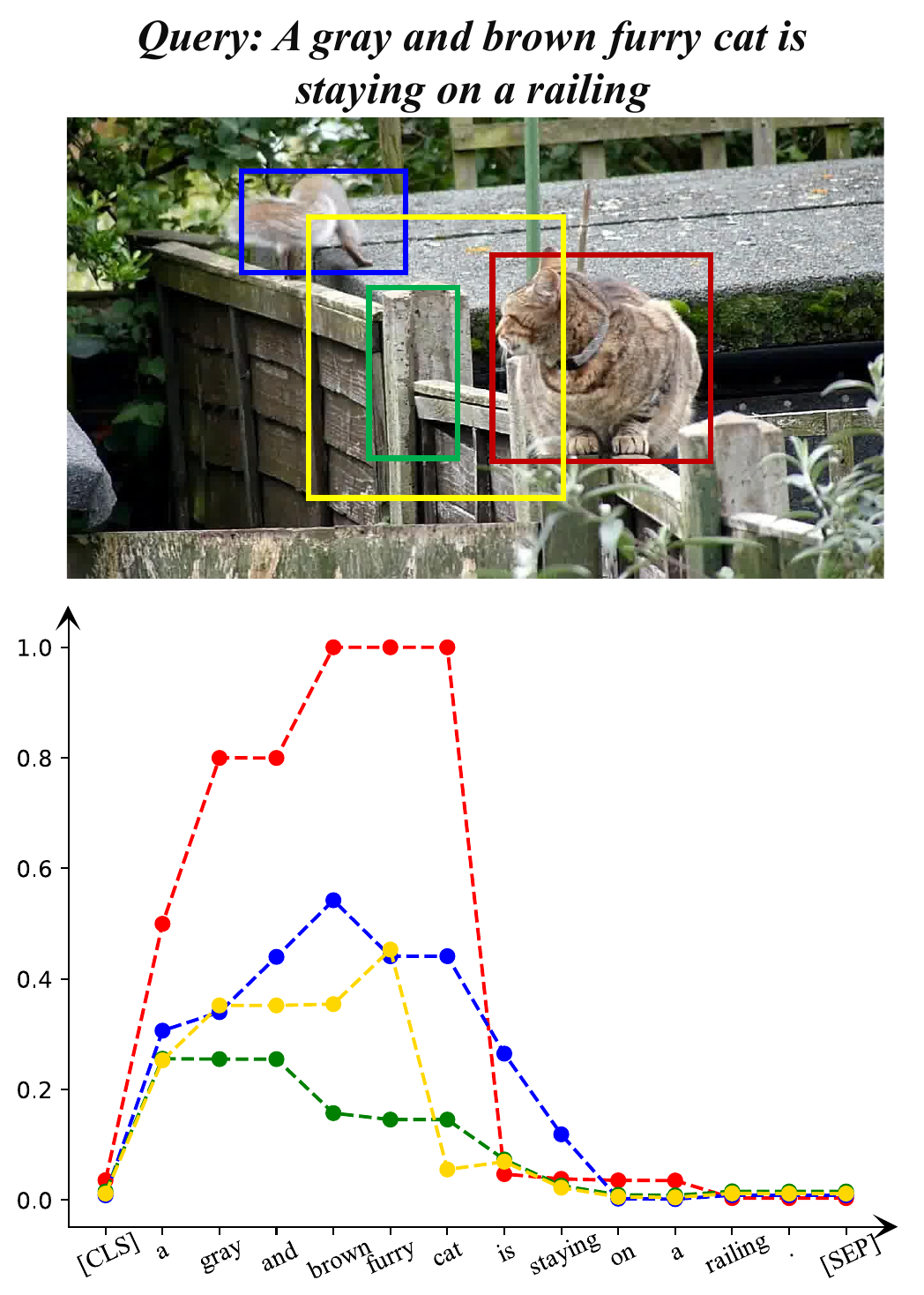}
	\caption{Illustration of fine-grained region-word level alignment. The regions marked by bounding boxes are corresponding to dashed lines according to color.}
	\label{fig:query_word}
\end{figure}

\noindent\textbf{Spatial resolution\footnote{We use ``resolution" to denote the long edge of a frame. \label{note:resolution}} and temporal length.}  As shown in Table~\ref{table:E_T_R_M}(b), we ablate how $T$ and the resolution affect the VID-Entity testing. We report the values of Accu.@0.5. Results in Table~\ref{table:E_T_R_M} show our model achieves the best performance on VID-Entity when setting $T=20$ and resolution to $672$. According to Table~\ref{table:E_T_R_M}(c), we find the saturated performance is achieved when setting $T=200$  and resolution to $224$ in VidSTG-Entity.

\begin{figure*}[t]
	\centering
	\includegraphics[width=0.88\textwidth]{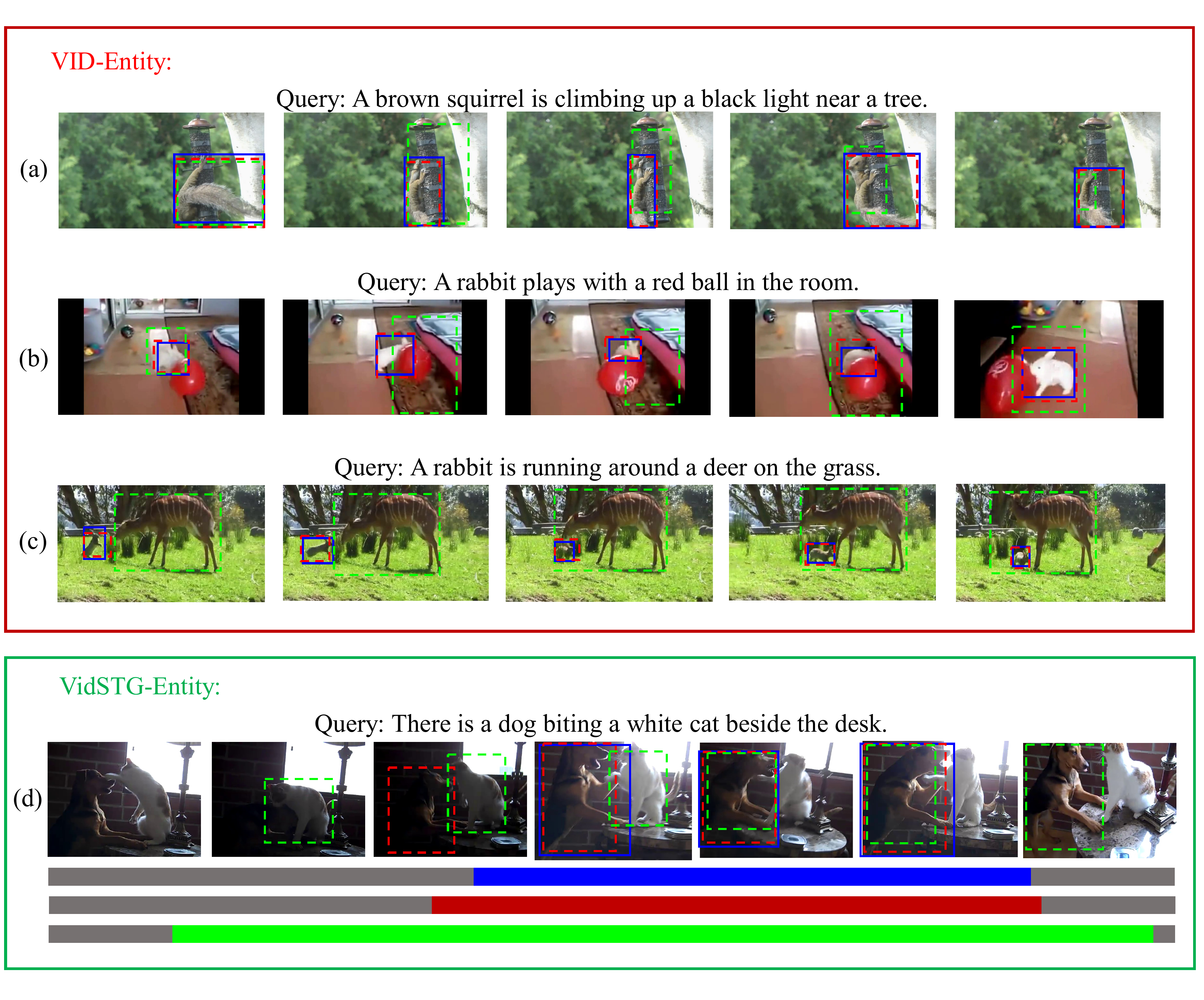}
	\vspace{-0.5em}
	\caption{Qualitative results of our ConFormer (marked in \textcolor{red}{red} dotted box), compared with \cite{song2021co} in VID-Sentence and \cite{yang2022tubedetr} in VidSTG. The compared methods are marked in \textcolor{green}{green} dotted box. The ground-truth annotations are marked in \textcolor{blue}{blue}.}
	\label{fig:visual_examples}
\end{figure*}

\begin{table}[t]
\caption{Different temporal modeling comparisons.}
\renewcommand\arraystretch{1.1}
\begin{center}
\resizebox{0.8\linewidth}{!}{
\begin{tabular}{cccccccccccccc}
\toprule
\textbf{Method} & Accu.@0.5 &m\_IoU \\

\midrule
ConFormer(\emph{wo/} Temporal) & 61.35 & 0.524   \\
ConFormer + Clip4Clip & 65.99 & 0.576   \\
ConFormer + Timesformer & \textbf{67.17} & \textbf{0.600}  \\
\bottomrule
\end{tabular}
}
\end{center}


\label{table:temporal modeling}
\end{table}
\noindent\textbf{Integration of visual and textual features.} To acquire better vision-language multimodal features, we explore integrating vision and language features. \emph{1) Concat:} A simple approach is to directly concatenate the video and text features into a sequence. \emph{2) Cross-Attention:} We fuse the video and text features with the vanilla cross-attention~\cite{vaswani2017attention}. Specifically, we treat the video feature as ``query" and the text feature as ``key"/``value". The results are shown in Table~\ref{table:fuse}, we find that simple concatenation can achieve a better performance in our framework.

\noindent\textbf{Temporal modeling.} To deal with the complex temporal information in the video, we explore some existing excellent temporal modeling modules. \emph{1)} CLIP4CLIP-seqTransf~\cite{luo2022clip4clip}
utilizes an extra Transofrmer Encoder to capture temporal information from patches of  a sequential of frames. 
\emph{2)} Timesformer~\cite{bertasius2021space} 
adapted the standard Transformer architecture, building exclusively on self-attention over space and time, to enable Transformer process temporal information.



\subsection{Visualization and Analysis}\label{sec:5.4}

\noindent\textbf{Region-word Alignment.} We visualize the fine-grained region-word alignment in Figure~\ref{fig:query_word} to demonstrate the effectiveness of our Entity-aware Contrastive Loss. Specifically, we select specific object queries and calculate their cosine similarity scores with each word in the described sentence. As shown in Figure~\ref{fig:query_word}, the query containing the target instance "\texttt{cat}" is highly corresponding to the word "\texttt{cat}", which manifests our motivation that Entity-aware Contrastive Loss leads to more fine-grained alignment. 

\noindent\textbf{Grounding results.}
We give the visualizations of our ConFormer and other excellent methods in Figure~\ref{fig:visual_examples}. Results demonstrate the superiority of ConFormer. Specifically, in Figure~\ref{fig:visual_examples}(a), our ConFormer localizes the region of "\texttt{squirrel}" precisely, while \cite{song2021co} fails to ground the target instance due to the interference of occlusion. In Figure~\ref{fig:visual_examples}(c), the target instance "\texttt{rabbit}" is too small and the word "\texttt{deer}" in the query sentence may confuse the model to localize the non-target instance. Nevertheless, our ConFormer still captures the overall semantic information and targets the corresponding instance. Similar to Figure~\ref{fig:visual_examples}(c), the example of VidSTG-Entity faces the same challenge in grounding target. However, ConFormer both localizes the target region precisely, showing the effectiveness of our Content-conditioned Query Generation and Entity-aware Contrastive Loss. 

\section{Conclusions}
We proposed ConFormer, a novel Transformer-based video REC method. Specifically, instead of using a fixed number of learnable embeddings designed for close-set categories, we propose a content-conditioned query generation module to generate dynamic queries . It efficiently addresses the open-end nature in video REC. Moreover, we propose an entity-aware contrastive loss to construct the fine-grained visual-text alignment for our ConFormer. Besides, we contributed two video datasets (\ie, VID-Entity and VidSTG-Entity) with region-phrase alignment annotations. Extensive results show that ConFormer achieves state-of-the-art performance on both (trimmed/untrimmed) video and image REC datasets.

\noindent \textbf{Acknowledgements.} This paper was partially supported by NSFC (No: 62176008) and Shenzhen Science \& Technology Research Program (No:GXWD20201231165807007-20200814115301001).

\balance
\bibliographystyle{ACM-Reference-Format}
\bibliography{acmart}


\begin{thebibliography}{73}


\ifx \showCODEN    \undefined \def \showCODEN     #1{\unskip}     \fi
\ifx \showDOI      \undefined \def \showDOI       #1{#1}\fi
\ifx \showISBNx    \undefined \def \showISBNx     #1{\unskip}     \fi
\ifx \showISBNxiii \undefined \def \showISBNxiii  #1{\unskip}     \fi
\ifx \showISSN     \undefined \def \showISSN      #1{\unskip}     \fi
\ifx \showLCCN     \undefined \def \showLCCN      #1{\unskip}     \fi
\ifx \shownote     \undefined \def \shownote      #1{#1}          \fi
\ifx \showarticletitle \undefined \def \showarticletitle #1{#1}   \fi
\ifx \showURL      \undefined \def \showURL       {\relax}        \fi
\providecommand\bibfield[2]{#2}
\providecommand\bibinfo[2]{#2}
\providecommand\natexlab[1]{#1}
\providecommand\showeprint[2][]{arXiv:#2}

\bibitem[Anderson et~al\mbox{.}(2018)]%
        {anderson2018bottom}
\bibfield{author}{\bibinfo{person}{Peter Anderson}, \bibinfo{person}{Xiaodong
  He}, \bibinfo{person}{Chris Buehler}, \bibinfo{person}{Damien Teney},
  \bibinfo{person}{Mark Johnson}, \bibinfo{person}{Stephen Gould}, {and}
  \bibinfo{person}{Lei Zhang}.} \bibinfo{year}{2018}\natexlab{}.
\newblock \showarticletitle{Bottom-up and top-down attention for image
  captioning and visual question answering}. In
  \bibinfo{booktitle}{\emph{Proceedings of the IEEE conference on computer
  vision and pattern recognition}}. \bibinfo{pages}{6077--6086}.
\newblock


\bibitem[Antol et~al\mbox{.}(2015)]%
        {antol2015vqa}
\bibfield{author}{\bibinfo{person}{Stanislaw Antol}, \bibinfo{person}{Aishwarya
  Agrawal}, \bibinfo{person}{Jiasen Lu}, \bibinfo{person}{Margaret Mitchell},
  \bibinfo{person}{Dhruv Batra}, \bibinfo{person}{C~Lawrence Zitnick}, {and}
  \bibinfo{person}{Devi Parikh}.} \bibinfo{year}{2015}\natexlab{}.
\newblock \showarticletitle{Vqa: Visual question answering}. In
  \bibinfo{booktitle}{\emph{Proceedings of the IEEE international conference on
  computer vision}}. \bibinfo{pages}{2425--2433}.
\newblock


\bibitem[Arandjelovic et~al\mbox{.}(2016)]%
        {arandjelovic2016netvlad}
\bibfield{author}{\bibinfo{person}{Relja Arandjelovic}, \bibinfo{person}{Petr
  Gronat}, \bibinfo{person}{Akihiko Torii}, \bibinfo{person}{Tomas Pajdla},
  {and} \bibinfo{person}{Josef Sivic}.} \bibinfo{year}{2016}\natexlab{}.
\newblock \showarticletitle{NetVLAD: CNN architecture for weakly supervised
  place recognition}. In \bibinfo{booktitle}{\emph{Proceedings of the IEEE
  conference on computer vision and pattern recognition}}.
  \bibinfo{pages}{5297--5307}.
\newblock


\bibitem[Cao et~al\mbox{.}(2021)]%
        {cao2021pursuit}
\bibfield{author}{\bibinfo{person}{Meng Cao}, \bibinfo{person}{Long Chen},
  \bibinfo{person}{Mike~Zheng Shou}, \bibinfo{person}{Can Zhang}, {and}
  \bibinfo{person}{Yuexian Zou}.} \bibinfo{year}{2021}\natexlab{}.
\newblock \showarticletitle{On Pursuit of Designing Multi-modal Transformer for
  Video Grounding}. In \bibinfo{booktitle}{\emph{Proceedings of the 2021
  Conference on Empirical Methods in Natural Language Processing}}.
  \bibinfo{pages}{9810--9823}.
\newblock


\bibitem[Cao et~al\mbox{.}(2022a)]%
        {cao2022correspondence}
\bibfield{author}{\bibinfo{person}{Meng Cao}, \bibinfo{person}{Ji Jiang},
  \bibinfo{person}{Long Chen}, {and} \bibinfo{person}{Yuexian Zou}.}
  \bibinfo{year}{2022}\natexlab{a}.
\newblock \showarticletitle{Correspondence matters for video referring
  expression comprehension}. In \bibinfo{booktitle}{\emph{Proceedings of the
  30th ACM International Conference on Multimedia}}.
  \bibinfo{pages}{4967--4976}.
\newblock


\bibitem[Cao et~al\mbox{.}(2023)]%
        {cao2023iterative}
\bibfield{author}{\bibinfo{person}{Meng Cao}, \bibinfo{person}{Fangyun Wei},
  \bibinfo{person}{Can Xu}, \bibinfo{person}{Xiubo Geng}, \bibinfo{person}{Long
  Chen}, \bibinfo{person}{Can Zhang}, \bibinfo{person}{Yuexian Zou},
  \bibinfo{person}{Tao Shen}, {and} \bibinfo{person}{Daxin Jiang}.}
  \bibinfo{year}{2023}\natexlab{}.
\newblock \showarticletitle{Iterative Proposal Refinement for Weakly-Supervised
  Video Grounding}. In \bibinfo{booktitle}{\emph{Proceedings of the IEEE/CVF
  Conference on Computer Vision and Pattern Recognition}}.
  \bibinfo{pages}{6524--6534}.
\newblock


\bibitem[Cao et~al\mbox{.}(2022b)]%
        {cao2022locvtp}
\bibfield{author}{\bibinfo{person}{Meng Cao}, \bibinfo{person}{Tianyu Yang},
  \bibinfo{person}{Junwu Weng}, \bibinfo{person}{Can Zhang},
  \bibinfo{person}{Jue Wang}, {and} \bibinfo{person}{Yuexian Zou}.}
  \bibinfo{year}{2022}\natexlab{b}.
\newblock \showarticletitle{Locvtp: Video-text pre-training for temporal
  localization}. In \bibinfo{booktitle}{\emph{European Conference on Computer
  Vision}}. Springer, \bibinfo{pages}{38--56}.
\newblock


\bibitem[Cao et~al\mbox{.}(2022c)]%
        {cao2022deep}
\bibfield{author}{\bibinfo{person}{Meng Cao}, \bibinfo{person}{Can Zhang},
  \bibinfo{person}{Long Chen}, \bibinfo{person}{Mike~Zheng Shou}, {and}
  \bibinfo{person}{Yuexian Zou}.} \bibinfo{year}{2022}\natexlab{c}.
\newblock \showarticletitle{Deep Motion Prior for Weakly-Supervised Temporal
  Action Localization}.
\newblock \bibinfo{journal}{\emph{IEEE Transactions on Image Processing}}
  \bibinfo{volume}{31} (\bibinfo{year}{2022}), \bibinfo{pages}{5203--5213}.
\newblock


\bibitem[Carion et~al\mbox{.}(2020)]%
        {carion2020end}
\bibfield{author}{\bibinfo{person}{Nicolas Carion}, \bibinfo{person}{Francisco
  Massa}, \bibinfo{person}{Gabriel Synnaeve}, \bibinfo{person}{Nicolas
  Usunier}, \bibinfo{person}{Alexander Kirillov}, {and} \bibinfo{person}{Sergey
  Zagoruyko}.} \bibinfo{year}{2020}\natexlab{}.
\newblock \showarticletitle{End-to-end object detection with transformers}. In
  \bibinfo{booktitle}{\emph{European conference on computer vision}}. Springer,
  \bibinfo{pages}{213--229}.
\newblock


\bibitem[Chen et~al\mbox{.}(2019a)]%
        {chen2019uniter}
\bibfield{author}{\bibinfo{person}{Yen-Chun Chen}, \bibinfo{person}{Linjie Li},
  \bibinfo{person}{Licheng Yu}, \bibinfo{person}{Ahmed El~Kholy},
  \bibinfo{person}{Faisal Ahmed}, \bibinfo{person}{Zhe Gan},
  \bibinfo{person}{Yu Cheng}, {and} \bibinfo{person}{Jingjing Liu}.}
  \bibinfo{year}{2019}\natexlab{a}.
\newblock \showarticletitle{Uniter: Learning universal image-text
  representations}.
\newblock  (\bibinfo{year}{2019}).
\newblock


\bibitem[Chen et~al\mbox{.}(2021)]%
        {chen2021end}
\bibfield{author}{\bibinfo{person}{Yi-Wen Chen}, \bibinfo{person}{Yi-Hsuan
  Tsai}, {and} \bibinfo{person}{Ming-Hsuan Yang}.}
  \bibinfo{year}{2021}\natexlab{}.
\newblock \showarticletitle{End-to-end multi-modal video temporal grounding}.
\newblock \bibinfo{journal}{\emph{Advances in Neural Information Processing
  Systems}}  \bibinfo{volume}{34} (\bibinfo{year}{2021}),
  \bibinfo{pages}{28442--28453}.
\newblock


\bibitem[Chen et~al\mbox{.}(2019b)]%
        {chen2019weakly}
\bibfield{author}{\bibinfo{person}{Zhenfang Chen}, \bibinfo{person}{Lin Ma},
  \bibinfo{person}{Wenhan Luo}, {and} \bibinfo{person}{Kwan-Yee~K Wong}.}
  \bibinfo{year}{2019}\natexlab{b}.
\newblock \showarticletitle{Weakly-supervised spatio-temporally grounding
  natural sentence in video}.
\newblock \bibinfo{journal}{\emph{arXiv preprint arXiv:1906.02549}}
  (\bibinfo{year}{2019}).
\newblock


\bibitem[Deng(2009)]%
        {deng2009large}
\bibfield{author}{\bibinfo{person}{Jia Deng}.} \bibinfo{year}{2009}\natexlab{}.
\newblock \showarticletitle{A large-scale hierarchical image database}.
\newblock \bibinfo{journal}{\emph{Proc. of IEEE Computer Vision and Pattern
  Recognition, 2009}} (\bibinfo{year}{2009}).
\newblock


\bibitem[Deng et~al\mbox{.}(2009)]%
        {deng2009imagenet}
\bibfield{author}{\bibinfo{person}{Jia Deng}, \bibinfo{person}{Wei Dong},
  \bibinfo{person}{Richard Socher}, \bibinfo{person}{Li-Jia Li},
  \bibinfo{person}{Kai Li}, {and} \bibinfo{person}{Li Fei-Fei}.}
  \bibinfo{year}{2009}\natexlab{}.
\newblock \showarticletitle{Imagenet: A large-scale hierarchical image
  database}. In \bibinfo{booktitle}{\emph{2009 IEEE conference on computer
  vision and pattern recognition}}. Ieee, \bibinfo{pages}{248--255}.
\newblock


\bibitem[Deng et~al\mbox{.}(2021)]%
        {deng2021transvg}
\bibfield{author}{\bibinfo{person}{Jiajun Deng}, \bibinfo{person}{Zhengyuan
  Yang}, \bibinfo{person}{Tianlang Chen}, \bibinfo{person}{Wengang Zhou}, {and}
  \bibinfo{person}{Houqiang Li}.} \bibinfo{year}{2021}\natexlab{}.
\newblock \showarticletitle{Transvg: End-to-end visual grounding with
  transformers}. In \bibinfo{booktitle}{\emph{Proceedings of the IEEE/CVF
  International Conference on Computer Vision}}. \bibinfo{pages}{1769--1779}.
\newblock


\bibitem[Dong et~al\mbox{.}(2019)]%
        {dong2019dual}
\bibfield{author}{\bibinfo{person}{Jianfeng Dong}, \bibinfo{person}{Xirong Li},
  \bibinfo{person}{Chaoxi Xu}, \bibinfo{person}{Shouling Ji},
  \bibinfo{person}{Yuan He}, \bibinfo{person}{Gang Yang}, {and}
  \bibinfo{person}{Xun Wang}.} \bibinfo{year}{2019}\natexlab{}.
\newblock \showarticletitle{Dual encoding for zero-example video retrieval}. In
  \bibinfo{booktitle}{\emph{Proceedings of the IEEE/CVF conference on computer
  vision and pattern recognition}}. \bibinfo{pages}{9346--9355}.
\newblock


\bibitem[Du et~al\mbox{.}(2022)]%
        {du2022visual}
\bibfield{author}{\bibinfo{person}{Ye Du}, \bibinfo{person}{Zehua Fu},
  \bibinfo{person}{Qingjie Liu}, {and} \bibinfo{person}{Yunhong Wang}.}
  \bibinfo{year}{2022}\natexlab{}.
\newblock \showarticletitle{Visual grounding with transformers}. In
  \bibinfo{booktitle}{\emph{2022 IEEE International Conference on Multimedia
  and Expo (ICME)}}. IEEE, \bibinfo{pages}{1--6}.
\newblock


\bibitem[\emph{et al.}([n.\,d.])]%
        {bertasius2021space}
\bibfield{author}{\bibinfo{person}{Bertasius \emph{et al.}}}
  \bibinfo{year}{[n.\,d.]}\natexlab{}.
\newblock \showarticletitle{Is space-time attention all you need for video
  understanding?}. In \bibinfo{booktitle}{\emph{ICML}}.
\newblock


\bibitem[Feng et~al\mbox{.}(2021a)]%
        {feng2021siamese}
\bibfield{author}{\bibinfo{person}{Qi Feng}, \bibinfo{person}{Vitaly Ablavsky},
  \bibinfo{person}{Qinxun Bai}, {and} \bibinfo{person}{Stan Sclaroff}.}
  \bibinfo{year}{2021}\natexlab{a}.
\newblock \showarticletitle{Siamese Natural Language Tracker: Tracking by
  Natural Language Descriptions with Siamese Trackers}. In
  \bibinfo{booktitle}{\emph{Proceedings of the IEEE/CVF Conference on Computer
  Vision and Pattern Recognition}}. \bibinfo{pages}{5851--5860}.
\newblock


\bibitem[Feng et~al\mbox{.}(2021b)]%
        {feng2021decoupled}
\bibfield{author}{\bibinfo{person}{Qianyu Feng}, \bibinfo{person}{Yunchao Wei},
  \bibinfo{person}{Mingming Cheng}, {and} \bibinfo{person}{Yi Yang}.}
  \bibinfo{year}{2021}\natexlab{b}.
\newblock \showarticletitle{Decoupled spatial temporal graphs for generic
  visual grounding}.
\newblock \bibinfo{journal}{\emph{arXiv preprint arXiv:2103.10191}}
  (\bibinfo{year}{2021}).
\newblock


\bibitem[Gan et~al\mbox{.}(2020)]%
        {gan2020large}
\bibfield{author}{\bibinfo{person}{Zhe Gan}, \bibinfo{person}{Yen-Chun Chen},
  \bibinfo{person}{Linjie Li}, \bibinfo{person}{Chen Zhu}, \bibinfo{person}{Yu
  Cheng}, {and} \bibinfo{person}{Jingjing Liu}.}
  \bibinfo{year}{2020}\natexlab{}.
\newblock \showarticletitle{Large-scale adversarial training for
  vision-and-language representation learning}.
\newblock \bibinfo{journal}{\emph{Advances in Neural Information Processing
  Systems}}  \bibinfo{volume}{33} (\bibinfo{year}{2020}),
  \bibinfo{pages}{6616--6628}.
\newblock


\bibitem[Gao et~al\mbox{.}(2017)]%
        {gao2017tall}
\bibfield{author}{\bibinfo{person}{Jiyang Gao}, \bibinfo{person}{Chen Sun},
  \bibinfo{person}{Zhenheng Yang}, {and} \bibinfo{person}{Ram Nevatia}.}
  \bibinfo{year}{2017}\natexlab{}.
\newblock \showarticletitle{Tall: Temporal activity localization via language
  query}. In \bibinfo{booktitle}{\emph{Proceedings of the IEEE international
  conference on computer vision}}. \bibinfo{pages}{5267--5275}.
\newblock


\bibitem[He et~al\mbox{.}(2016)]%
        {he2016deep}
\bibfield{author}{\bibinfo{person}{Kaiming He}, \bibinfo{person}{Xiangyu
  Zhang}, \bibinfo{person}{Shaoqing Ren}, {and} \bibinfo{person}{Jian Sun}.}
  \bibinfo{year}{2016}\natexlab{}.
\newblock \showarticletitle{Deep residual learning for image recognition}. In
  \bibinfo{booktitle}{\emph{Proceedings of the IEEE conference on computer
  vision and pattern recognition}}. \bibinfo{pages}{770--778}.
\newblock


\bibitem[Honnibal and Johnson(2015)]%
        {honnibal2015improved}
\bibfield{author}{\bibinfo{person}{Matthew Honnibal} {and}
  \bibinfo{person}{Mark Johnson}.} \bibinfo{year}{2015}\natexlab{}.
\newblock \showarticletitle{An improved non-monotonic transition system for
  dependency parsing}. In \bibinfo{booktitle}{\emph{Proceedings of the 2015
  conference on empirical methods in natural language processing}}.
  \bibinfo{pages}{1373--1378}.
\newblock


\bibitem[Hu et~al\mbox{.}(2017)]%
        {hu2017modeling}
\bibfield{author}{\bibinfo{person}{Ronghang Hu}, \bibinfo{person}{Marcus
  Rohrbach}, \bibinfo{person}{Jacob Andreas}, \bibinfo{person}{Trevor Darrell},
  {and} \bibinfo{person}{Kate Saenko}.} \bibinfo{year}{2017}\natexlab{}.
\newblock \showarticletitle{Modeling relationships in referential expressions
  with compositional modular networks}. In
  \bibinfo{booktitle}{\emph{Proceedings of the IEEE conference on computer
  vision and pattern recognition}}. \bibinfo{pages}{1115--1124}.
\newblock


\bibitem[Hu et~al\mbox{.}(2016)]%
        {hu2016natural}
\bibfield{author}{\bibinfo{person}{Ronghang Hu}, \bibinfo{person}{Huazhe Xu},
  \bibinfo{person}{Marcus Rohrbach}, \bibinfo{person}{Jiashi Feng},
  \bibinfo{person}{Kate Saenko}, {and} \bibinfo{person}{Trevor Darrell}.}
  \bibinfo{year}{2016}\natexlab{}.
\newblock \showarticletitle{Natural language object retrieval}. In
  \bibinfo{booktitle}{\emph{Proceedings of the IEEE conference on computer
  vision and pattern recognition}}. \bibinfo{pages}{4555--4564}.
\newblock


\bibitem[Huang et~al\mbox{.}(2018)]%
        {huang2018finding}
\bibfield{author}{\bibinfo{person}{De-An Huang}, \bibinfo{person}{Shyamal
  Buch}, \bibinfo{person}{Lucio Dery}, \bibinfo{person}{Animesh Garg},
  \bibinfo{person}{Li Fei-Fei}, {and} \bibinfo{person}{Juan~Carlos Niebles}.}
  \bibinfo{year}{2018}\natexlab{}.
\newblock \showarticletitle{Finding" it": Weakly-supervised reference-aware
  visual grounding in instructional videos}. In
  \bibinfo{booktitle}{\emph{Proceedings of the IEEE Conference on Computer
  Vision and Pattern Recognition}}. \bibinfo{pages}{5948--5957}.
\newblock


\bibitem[Kamath et~al\mbox{.}(2021)]%
        {kamath2021mdetr}
\bibfield{author}{\bibinfo{person}{Aishwarya Kamath}, \bibinfo{person}{Mannat
  Singh}, \bibinfo{person}{Yann LeCun}, \bibinfo{person}{Gabriel Synnaeve},
  \bibinfo{person}{Ishan Misra}, {and} \bibinfo{person}{Nicolas Carion}.}
  \bibinfo{year}{2021}\natexlab{}.
\newblock \showarticletitle{MDETR-modulated detection for end-to-end
  multi-modal understanding}. In \bibinfo{booktitle}{\emph{Proceedings of the
  IEEE/CVF International Conference on Computer Vision}}.
  \bibinfo{pages}{1780--1790}.
\newblock


\bibitem[Karpathy and Fei-Fei(2015)]%
        {karpathy2015deep}
\bibfield{author}{\bibinfo{person}{Andrej Karpathy} {and} \bibinfo{person}{Li
  Fei-Fei}.} \bibinfo{year}{2015}\natexlab{}.
\newblock \showarticletitle{Deep visual-semantic alignments for generating
  image descriptions}. In \bibinfo{booktitle}{\emph{Proceedings of the IEEE
  conference on computer vision and pattern recognition}}.
  \bibinfo{pages}{3128--3137}.
\newblock


\bibitem[Krishna et~al\mbox{.}(2017)]%
        {krishna2017visual}
\bibfield{author}{\bibinfo{person}{Ranjay Krishna}, \bibinfo{person}{Yuke Zhu},
  \bibinfo{person}{Oliver Groth}, \bibinfo{person}{Justin Johnson},
  \bibinfo{person}{Kenji Hata}, \bibinfo{person}{Joshua Kravitz},
  \bibinfo{person}{Stephanie Chen}, \bibinfo{person}{Yannis Kalantidis},
  \bibinfo{person}{Li-Jia Li}, \bibinfo{person}{David~A Shamma},
  {et~al\mbox{.}}} \bibinfo{year}{2017}\natexlab{}.
\newblock \showarticletitle{Visual genome: Connecting language and vision using
  crowdsourced dense image annotations}.
\newblock \bibinfo{journal}{\emph{International journal of computer vision}}
  \bibinfo{volume}{123}, \bibinfo{number}{1} (\bibinfo{year}{2017}),
  \bibinfo{pages}{32--73}.
\newblock


\bibitem[Kuhn(1955)]%
        {kuhn1955hungarian}
\bibfield{author}{\bibinfo{person}{Harold~W Kuhn}.}
  \bibinfo{year}{1955}\natexlab{}.
\newblock \showarticletitle{The Hungarian method for the assignment problem}.
\newblock \bibinfo{journal}{\emph{Naval research logistics quarterly}}
  \bibinfo{volume}{2}, \bibinfo{number}{1-2} (\bibinfo{year}{1955}),
  \bibinfo{pages}{83--97}.
\newblock


\bibitem[Li et~al\mbox{.}(2019)]%
        {li2019siamrpn++}
\bibfield{author}{\bibinfo{person}{Bo Li}, \bibinfo{person}{Wei Wu},
  \bibinfo{person}{Qiang Wang}, \bibinfo{person}{Fangyi Zhang},
  \bibinfo{person}{Junliang Xing}, {and} \bibinfo{person}{Junjie Yan}.}
  \bibinfo{year}{2019}\natexlab{}.
\newblock \showarticletitle{Siamrpn++: Evolution of siamese visual tracking
  with very deep networks}. In \bibinfo{booktitle}{\emph{Proceedings of the
  IEEE/CVF Conference on Computer Vision and Pattern Recognition}}.
  \bibinfo{pages}{4282--4291}.
\newblock


\bibitem[Li et~al\mbox{.}(2023a)]%
        {li2023g2l}
\bibfield{author}{\bibinfo{person}{Hongxiang Li}, \bibinfo{person}{Meng Cao},
  \bibinfo{person}{Xuxin Cheng}, \bibinfo{person}{Yaowei Li},
  \bibinfo{person}{Zhihong Zhu}, {and} \bibinfo{person}{Yuexian Zou}.}
  \bibinfo{year}{2023}\natexlab{a}.
\newblock \showarticletitle{G2L: Semantically Aligned and Uniform Video
  Grounding via Geodesic and Game Theory}.
\newblock \bibinfo{journal}{\emph{arXiv preprint arXiv:2307.14277}}
  (\bibinfo{year}{2023}).
\newblock


\bibitem[Li et~al\mbox{.}(2023b)]%
        {li2023generating}
\bibfield{author}{\bibinfo{person}{Hongxiang Li}, \bibinfo{person}{Meng Cao},
  \bibinfo{person}{Xuxin Cheng}, \bibinfo{person}{Zhihong Zhu},
  \bibinfo{person}{Yaowei Li}, {and} \bibinfo{person}{Yuexian Zou}.}
  \bibinfo{year}{2023}\natexlab{b}.
\newblock \showarticletitle{Generating templated caption for video grounding}.
\newblock \bibinfo{journal}{\emph{arXiv preprint arXiv:2301.05997}}
  (\bibinfo{year}{2023}).
\newblock


\bibitem[Li and Sigal(2021)]%
        {li2021referring}
\bibfield{author}{\bibinfo{person}{Muchen Li} {and} \bibinfo{person}{Leonid
  Sigal}.} \bibinfo{year}{2021}\natexlab{}.
\newblock \showarticletitle{Referring transformer: A one-step approach to
  multi-task visual grounding}.
\newblock \bibinfo{journal}{\emph{Advances in neural information processing
  systems}}  \bibinfo{volume}{34} (\bibinfo{year}{2021}),
  \bibinfo{pages}{19652--19664}.
\newblock


\bibitem[Lin et~al\mbox{.}(2014)]%
        {lin2014microsoft}
\bibfield{author}{\bibinfo{person}{Tsung-Yi Lin}, \bibinfo{person}{Michael
  Maire}, \bibinfo{person}{Serge Belongie}, \bibinfo{person}{James Hays},
  \bibinfo{person}{Pietro Perona}, \bibinfo{person}{Deva Ramanan},
  \bibinfo{person}{Piotr Doll{\'a}r}, {and} \bibinfo{person}{C~Lawrence
  Zitnick}.} \bibinfo{year}{2014}\natexlab{}.
\newblock \showarticletitle{Microsoft coco: Common objects in context}. In
  \bibinfo{booktitle}{\emph{Computer Vision--ECCV 2014: 13th European
  Conference, Zurich, Switzerland, September 6-12, 2014, Proceedings, Part V
  13}}. Springer, \bibinfo{pages}{740--755}.
\newblock


\bibitem[Liu et~al\mbox{.}(2019b)]%
        {liu2019learning}
\bibfield{author}{\bibinfo{person}{Daqing Liu}, \bibinfo{person}{Hanwang
  Zhang}, \bibinfo{person}{Feng Wu}, {and} \bibinfo{person}{Zheng-Jun Zha}.}
  \bibinfo{year}{2019}\natexlab{b}.
\newblock \showarticletitle{Learning to assemble neural module tree networks
  for visual grounding}. In \bibinfo{booktitle}{\emph{Proceedings of the
  IEEE/CVF International Conference on Computer Vision}}.
  \bibinfo{pages}{4673--4682}.
\newblock


\bibitem[Liu et~al\mbox{.}(2022)]%
        {liu2022dab}
\bibfield{author}{\bibinfo{person}{Shilong Liu}, \bibinfo{person}{Feng Li},
  \bibinfo{person}{Hao Zhang}, \bibinfo{person}{Xiao Yang},
  \bibinfo{person}{Xianbiao Qi}, \bibinfo{person}{Hang Su},
  \bibinfo{person}{Jun Zhu}, {and} \bibinfo{person}{Lei Zhang}.}
  \bibinfo{year}{2022}\natexlab{}.
\newblock \showarticletitle{DAB-DETR: Dynamic anchor boxes are better queries
  for DETR}.
\newblock \bibinfo{journal}{\emph{arXiv preprint arXiv:2201.12329}}
  (\bibinfo{year}{2022}).
\newblock


\bibitem[Liu et~al\mbox{.}(2019a)]%
        {liu2019roberta}
\bibfield{author}{\bibinfo{person}{Yinhan Liu}, \bibinfo{person}{Myle Ott},
  \bibinfo{person}{Naman Goyal}, \bibinfo{person}{Jingfei Du},
  \bibinfo{person}{Mandar Joshi}, \bibinfo{person}{Danqi Chen},
  \bibinfo{person}{Omer Levy}, \bibinfo{person}{Mike Lewis},
  \bibinfo{person}{Luke Zettlemoyer}, {and} \bibinfo{person}{Veselin
  Stoyanov}.} \bibinfo{year}{2019}\natexlab{a}.
\newblock \showarticletitle{Roberta: A robustly optimized bert pretraining
  approach}.
\newblock \bibinfo{journal}{\emph{arXiv preprint arXiv:1907.11692}}
  (\bibinfo{year}{2019}).
\newblock


\bibitem[Loshchilov and Hutter(2017)]%
        {loshchilov2017decoupled}
\bibfield{author}{\bibinfo{person}{Ilya Loshchilov} {and}
  \bibinfo{person}{Frank Hutter}.} \bibinfo{year}{2017}\natexlab{}.
\newblock \showarticletitle{Decoupled weight decay regularization}.
\newblock \bibinfo{journal}{\emph{arXiv preprint arXiv:1711.05101}}
  (\bibinfo{year}{2017}).
\newblock


\bibitem[Luo([n.\,d.])]%
        {luo2022clip4clip}
\bibfield{author}{\bibinfo{person}{Huaishao~\emph{et al.} Luo}.}
  \bibinfo{year}{[n.\,d.]}\natexlab{}.
\newblock \showarticletitle{CLIP4Clip: An empirical study of CLIP for end to
  end video clip retrieval and captioning}.
\newblock  (\bibinfo{year}{[n.\,d.]}).
\newblock


\bibitem[Mao et~al\mbox{.}(2023)]%
        {mao2023improving}
\bibfield{author}{\bibinfo{person}{Yangjun Mao}, \bibinfo{person}{Jun Xiao},
  \bibinfo{person}{Dong Zhang}, \bibinfo{person}{Meng Cao},
  \bibinfo{person}{Jian Shao}, \bibinfo{person}{Yueting Zhuang}, {and}
  \bibinfo{person}{Long Chen}.} \bibinfo{year}{2023}\natexlab{}.
\newblock \showarticletitle{Improving Reference-based Distinctive Image
  Captioning with Contrastive Rewards}.
\newblock \bibinfo{journal}{\emph{arXiv preprint arXiv:2306.14259}}
  (\bibinfo{year}{2023}).
\newblock


\bibitem[Meng et~al\mbox{.}(2021)]%
        {meng2021conditional}
\bibfield{author}{\bibinfo{person}{Depu Meng}, \bibinfo{person}{Xiaokang Chen},
  \bibinfo{person}{Zejia Fan}, \bibinfo{person}{Gang Zeng},
  \bibinfo{person}{Houqiang Li}, \bibinfo{person}{Yuhui Yuan},
  \bibinfo{person}{Lei Sun}, {and} \bibinfo{person}{Jingdong Wang}.}
  \bibinfo{year}{2021}\natexlab{}.
\newblock \showarticletitle{Conditional detr for fast training convergence}. In
  \bibinfo{booktitle}{\emph{Proceedings of the IEEE/CVF International
  Conference on Computer Vision}}. \bibinfo{pages}{3651--3660}.
\newblock


\bibitem[Miech et~al\mbox{.}(2020)]%
        {miech2020end}
\bibfield{author}{\bibinfo{person}{Antoine Miech},
  \bibinfo{person}{Jean-Baptiste Alayrac}, \bibinfo{person}{Lucas Smaira},
  \bibinfo{person}{Ivan Laptev}, \bibinfo{person}{Josef Sivic}, {and}
  \bibinfo{person}{Andrew Zisserman}.} \bibinfo{year}{2020}\natexlab{}.
\newblock \showarticletitle{End-to-end learning of visual representations from
  uncurated instructional videos}. In \bibinfo{booktitle}{\emph{Proceedings of
  the IEEE/CVF Conference on Computer Vision and Pattern Recognition}}.
  \bibinfo{pages}{9879--9889}.
\newblock


\bibitem[Plummer et~al\mbox{.}(2015)]%
        {plummer2015flickr30k}
\bibfield{author}{\bibinfo{person}{Bryan~A Plummer}, \bibinfo{person}{Liwei
  Wang}, \bibinfo{person}{Chris~M Cervantes}, \bibinfo{person}{Juan~C Caicedo},
  \bibinfo{person}{Julia Hockenmaier}, {and} \bibinfo{person}{Svetlana
  Lazebnik}.} \bibinfo{year}{2015}\natexlab{}.
\newblock \showarticletitle{Flickr30k entities: Collecting region-to-phrase
  correspondences for richer image-to-sentence models}. In
  \bibinfo{booktitle}{\emph{Proceedings of the IEEE international conference on
  computer vision}}. \bibinfo{pages}{2641--2649}.
\newblock


\bibitem[Radford et~al\mbox{.}(2021)]%
        {radford2021learning}
\bibfield{author}{\bibinfo{person}{Alec Radford}, \bibinfo{person}{Jong~Wook
  Kim}, \bibinfo{person}{Chris Hallacy}, \bibinfo{person}{Aditya Ramesh},
  \bibinfo{person}{Gabriel Goh}, \bibinfo{person}{Sandhini Agarwal},
  \bibinfo{person}{Girish Sastry}, \bibinfo{person}{Amanda Askell},
  \bibinfo{person}{Pamela Mishkin}, \bibinfo{person}{Jack Clark},
  {et~al\mbox{.}}} \bibinfo{year}{2021}\natexlab{}.
\newblock \showarticletitle{Learning transferable visual models from natural
  language supervision}. In \bibinfo{booktitle}{\emph{International Conference
  on Machine Learning}}. PMLR, \bibinfo{pages}{8748--8763}.
\newblock


\bibitem[Rezatofighi et~al\mbox{.}(2019)]%
        {rezatofighi2019generalized}
\bibfield{author}{\bibinfo{person}{Hamid Rezatofighi}, \bibinfo{person}{Nathan
  Tsoi}, \bibinfo{person}{JunYoung Gwak}, \bibinfo{person}{Amir Sadeghian},
  \bibinfo{person}{Ian Reid}, {and} \bibinfo{person}{Silvio Savarese}.}
  \bibinfo{year}{2019}\natexlab{}.
\newblock \showarticletitle{Generalized intersection over union: A metric and a
  loss for bounding box regression}. In \bibinfo{booktitle}{\emph{Proceedings
  of the IEEE/CVF conference on computer vision and pattern recognition}}.
  \bibinfo{pages}{658--666}.
\newblock


\bibitem[Rohrbach et~al\mbox{.}(2016)]%
        {rohrbach2016grounding}
\bibfield{author}{\bibinfo{person}{Anna Rohrbach}, \bibinfo{person}{Marcus
  Rohrbach}, \bibinfo{person}{Ronghang Hu}, \bibinfo{person}{Trevor Darrell},
  {and} \bibinfo{person}{Bernt Schiele}.} \bibinfo{year}{2016}\natexlab{}.
\newblock \showarticletitle{Grounding of textual phrases in images by
  reconstruction}. In \bibinfo{booktitle}{\emph{European Conference on Computer
  Vision}}. Springer, \bibinfo{pages}{817--834}.
\newblock


\bibitem[Sadhu et~al\mbox{.}(2020)]%
        {sadhu2020video}
\bibfield{author}{\bibinfo{person}{Arka Sadhu}, \bibinfo{person}{Kan Chen},
  {and} \bibinfo{person}{Ram Nevatia}.} \bibinfo{year}{2020}\natexlab{}.
\newblock \showarticletitle{Video object grounding using semantic roles in
  language description}. In \bibinfo{booktitle}{\emph{Proceedings of the
  IEEE/CVF Conference on Computer Vision and Pattern Recognition}}.
  \bibinfo{pages}{10417--10427}.
\newblock


\bibitem[Sharma et~al\mbox{.}(2018)]%
        {sharma2018conceptual}
\bibfield{author}{\bibinfo{person}{Piyush Sharma}, \bibinfo{person}{Nan Ding},
  \bibinfo{person}{Sebastian Goodman}, {and} \bibinfo{person}{Radu Soricut}.}
  \bibinfo{year}{2018}\natexlab{}.
\newblock \showarticletitle{Conceptual captions: A cleaned, hypernymed, image
  alt-text dataset for automatic image captioning}. In
  \bibinfo{booktitle}{\emph{Proceedings of the 56th Annual Meeting of the
  Association for Computational Linguistics (Volume 1: Long Papers)}}.
  \bibinfo{pages}{2556--2565}.
\newblock


\bibitem[Song et~al\mbox{.}(2021)]%
        {song2021co}
\bibfield{author}{\bibinfo{person}{Sijie Song}, \bibinfo{person}{Xudong Lin},
  \bibinfo{person}{Jiaying Liu}, \bibinfo{person}{Zongming Guo}, {and}
  \bibinfo{person}{Shih-Fu Chang}.} \bibinfo{year}{2021}\natexlab{}.
\newblock \showarticletitle{Co-Grounding Networks with Semantic Attention for
  Referring Expression Comprehension in Videos}. In
  \bibinfo{booktitle}{\emph{Proceedings of the IEEE/CVF Conference on Computer
  Vision and Pattern Recognition}}. \bibinfo{pages}{1346--1355}.
\newblock


\bibitem[Su et~al\mbox{.}(2021)]%
        {su2021stvgbert}
\bibfield{author}{\bibinfo{person}{Rui Su}, \bibinfo{person}{Qian Yu}, {and}
  \bibinfo{person}{Dong Xu}.} \bibinfo{year}{2021}\natexlab{}.
\newblock \showarticletitle{Stvgbert: A visual-linguistic transformer based
  framework for spatio-temporal video grounding}. In
  \bibinfo{booktitle}{\emph{Proceedings of the IEEE/CVF International
  Conference on Computer Vision}}. \bibinfo{pages}{1533--1542}.
\newblock


\bibitem[Su et~al\mbox{.}(2019)]%
        {su2019vl}
\bibfield{author}{\bibinfo{person}{Weijie Su}, \bibinfo{person}{Xizhou Zhu},
  \bibinfo{person}{Yue Cao}, \bibinfo{person}{Bin Li}, \bibinfo{person}{Lewei
  Lu}, \bibinfo{person}{Furu Wei}, {and} \bibinfo{person}{Jifeng Dai}.}
  \bibinfo{year}{2019}\natexlab{}.
\newblock \showarticletitle{Vl-bert: Pre-training of generic visual-linguistic
  representations}.
\newblock \bibinfo{journal}{\emph{arXiv preprint arXiv:1908.08530}}
  (\bibinfo{year}{2019}).
\newblock


\bibitem[Vasudevan et~al\mbox{.}(2018)]%
        {vasudevan2018object}
\bibfield{author}{\bibinfo{person}{Arun~Balajee Vasudevan},
  \bibinfo{person}{Dengxin Dai}, {and} \bibinfo{person}{Luc Van~Gool}.}
  \bibinfo{year}{2018}\natexlab{}.
\newblock \showarticletitle{Object referring in videos with language and human
  gaze}. In \bibinfo{booktitle}{\emph{Proceedings of the IEEE Conference on
  Computer Vision and Pattern Recognition}}. \bibinfo{pages}{4129--4138}.
\newblock


\bibitem[Vaswani et~al\mbox{.}(2017)]%
        {vaswani2017attention}
\bibfield{author}{\bibinfo{person}{Ashish Vaswani}, \bibinfo{person}{Noam
  Shazeer}, \bibinfo{person}{Niki Parmar}, \bibinfo{person}{Jakob Uszkoreit},
  \bibinfo{person}{Llion Jones}, \bibinfo{person}{Aidan~N Gomez},
  \bibinfo{person}{{\L}ukasz Kaiser}, {and} \bibinfo{person}{Illia
  Polosukhin}.} \bibinfo{year}{2017}\natexlab{}.
\newblock \showarticletitle{Attention is all you need}.
\newblock \bibinfo{journal}{\emph{Advances in neural information processing
  systems}}  \bibinfo{volume}{30} (\bibinfo{year}{2017}).
\newblock


\bibitem[Wang et~al\mbox{.}(2021)]%
        {wang2021anchor}
\bibfield{author}{\bibinfo{person}{Yingming Wang}, \bibinfo{person}{Xiangyu
  Zhang}, \bibinfo{person}{Tong Yang}, {and} \bibinfo{person}{Jian Sun}.}
  \bibinfo{year}{2021}\natexlab{}.
\newblock \showarticletitle{Anchor detr: Query design for transformer-based
  detector}.
\newblock \bibinfo{journal}{\emph{arXiv preprint arXiv:2109.07107}}
  (\bibinfo{year}{2021}).
\newblock


\bibitem[Wolf et~al\mbox{.}(2019)]%
        {wolf2019huggingface}
\bibfield{author}{\bibinfo{person}{Thomas Wolf}, \bibinfo{person}{Lysandre
  Debut}, \bibinfo{person}{Victor Sanh}, \bibinfo{person}{Julien Chaumond},
  \bibinfo{person}{Clement Delangue}, \bibinfo{person}{Anthony Moi},
  \bibinfo{person}{Pierric Cistac}, \bibinfo{person}{Tim Rault},
  \bibinfo{person}{R{\'e}mi Louf}, \bibinfo{person}{Morgan Funtowicz},
  {et~al\mbox{.}}} \bibinfo{year}{2019}\natexlab{}.
\newblock \showarticletitle{Huggingface's transformers: State-of-the-art
  natural language processing}.
\newblock \bibinfo{journal}{\emph{arXiv preprint arXiv:1910.03771}}
  (\bibinfo{year}{2019}).
\newblock


\bibitem[Yang et~al\mbox{.}(2022)]%
        {yang2022tubedetr}
\bibfield{author}{\bibinfo{person}{Antoine Yang}, \bibinfo{person}{Antoine
  Miech}, \bibinfo{person}{Josef Sivic}, \bibinfo{person}{Ivan Laptev}, {and}
  \bibinfo{person}{Cordelia Schmid}.} \bibinfo{year}{2022}\natexlab{}.
\newblock \showarticletitle{TubeDETR: Spatio-Temporal Video Grounding with
  Transformers}.
\newblock \bibinfo{journal}{\emph{arXiv preprint arXiv:2203.16434}}
  (\bibinfo{year}{2022}).
\newblock


\bibitem[Yang et~al\mbox{.}(2019b)]%
        {yang2019dynamic}
\bibfield{author}{\bibinfo{person}{Sibei Yang}, \bibinfo{person}{Guanbin Li},
  {and} \bibinfo{person}{Yizhou Yu}.} \bibinfo{year}{2019}\natexlab{b}.
\newblock \showarticletitle{Dynamic graph attention for referring expression
  comprehension}. In \bibinfo{booktitle}{\emph{Proceedings of the IEEE/CVF
  International Conference on Computer Vision}}. \bibinfo{pages}{4644--4653}.
\newblock


\bibitem[Yang et~al\mbox{.}(2020)]%
        {yang2020improving}
\bibfield{author}{\bibinfo{person}{Zhengyuan Yang}, \bibinfo{person}{Tianlang
  Chen}, \bibinfo{person}{Liwei Wang}, {and} \bibinfo{person}{Jiebo Luo}.}
  \bibinfo{year}{2020}\natexlab{}.
\newblock \showarticletitle{Improving one-stage visual grounding by recursive
  sub-query construction}. In \bibinfo{booktitle}{\emph{European Conference on
  Computer Vision}}. Springer, \bibinfo{pages}{387--404}.
\newblock


\bibitem[Yang et~al\mbox{.}(2019a)]%
        {yang2019fast}
\bibfield{author}{\bibinfo{person}{Zhengyuan Yang}, \bibinfo{person}{Boqing
  Gong}, \bibinfo{person}{Liwei Wang}, \bibinfo{person}{Wenbing Huang},
  \bibinfo{person}{Dong Yu}, {and} \bibinfo{person}{Jiebo Luo}.}
  \bibinfo{year}{2019}\natexlab{a}.
\newblock \showarticletitle{A fast and accurate one-stage approach to visual
  grounding}. In \bibinfo{booktitle}{\emph{Proceedings of the IEEE/CVF
  International Conference on Computer Vision}}. \bibinfo{pages}{4683--4693}.
\newblock


\bibitem[Yu et~al\mbox{.}(2018)]%
        {yu2018mattnet}
\bibfield{author}{\bibinfo{person}{Licheng Yu}, \bibinfo{person}{Zhe Lin},
  \bibinfo{person}{Xiaohui Shen}, \bibinfo{person}{Jimei Yang},
  \bibinfo{person}{Xin Lu}, \bibinfo{person}{Mohit Bansal}, {and}
  \bibinfo{person}{Tamara~L Berg}.} \bibinfo{year}{2018}\natexlab{}.
\newblock \showarticletitle{Mattnet: Modular attention network for referring
  expression comprehension}. In \bibinfo{booktitle}{\emph{Proceedings of the
  IEEE Conference on Computer Vision and Pattern Recognition}}.
  \bibinfo{pages}{1307--1315}.
\newblock


\bibitem[Yu et~al\mbox{.}(2016)]%
        {yu2016modeling}
\bibfield{author}{\bibinfo{person}{Licheng Yu}, \bibinfo{person}{Patrick
  Poirson}, \bibinfo{person}{Shan Yang}, \bibinfo{person}{Alexander~C Berg},
  {and} \bibinfo{person}{Tamara~L Berg}.} \bibinfo{year}{2016}\natexlab{}.
\newblock \showarticletitle{Modeling context in referring expressions}. In
  \bibinfo{booktitle}{\emph{European Conference on Computer Vision}}. Springer,
  \bibinfo{pages}{69--85}.
\newblock


\bibitem[Yu et~al\mbox{.}(2017)]%
        {yu2017joint}
\bibfield{author}{\bibinfo{person}{Licheng Yu}, \bibinfo{person}{Hao Tan},
  \bibinfo{person}{Mohit Bansal}, {and} \bibinfo{person}{Tamara~L Berg}.}
  \bibinfo{year}{2017}\natexlab{}.
\newblock \showarticletitle{A joint speaker-listener-reinforcer model for
  referring expressions}. In \bibinfo{booktitle}{\emph{Proceedings of the IEEE
  Conference on Computer Vision and Pattern Recognition}}.
  \bibinfo{pages}{7282--7290}.
\newblock


\bibitem[Zeng et~al\mbox{.}(2020)]%
        {zeng2020dense}
\bibfield{author}{\bibinfo{person}{Runhao Zeng}, \bibinfo{person}{Haoming Xu},
  \bibinfo{person}{Wenbing Huang}, \bibinfo{person}{Peihao Chen},
  \bibinfo{person}{Mingkui Tan}, {and} \bibinfo{person}{Chuang Gan}.}
  \bibinfo{year}{2020}\natexlab{}.
\newblock \showarticletitle{Dense regression network for video grounding}. In
  \bibinfo{booktitle}{\emph{Proceedings of the IEEE/CVF Conference on Computer
  Vision and Pattern Recognition}}. \bibinfo{pages}{10287--10296}.
\newblock


\bibitem[Zhang et~al\mbox{.}(2021)]%
        {zhang2021cola}
\bibfield{author}{\bibinfo{person}{Can Zhang}, \bibinfo{person}{Meng Cao},
  \bibinfo{person}{Dongming Yang}, \bibinfo{person}{Jie Chen}, {and}
  \bibinfo{person}{Yuexian Zou}.} \bibinfo{year}{2021}\natexlab{}.
\newblock \showarticletitle{Cola: Weakly-supervised temporal action
  localization with snippet contrastive learning}. In
  \bibinfo{booktitle}{\emph{Proceedings of the IEEE/CVF Conference on Computer
  Vision and Pattern Recognition}}. \bibinfo{pages}{16010--16019}.
\newblock


\bibitem[Zhang et~al\mbox{.}(2022)]%
        {zhang2022unsupervised}
\bibfield{author}{\bibinfo{person}{Can Zhang}, \bibinfo{person}{Tianyu Yang},
  \bibinfo{person}{Junwu Weng}, \bibinfo{person}{Meng Cao},
  \bibinfo{person}{Jue Wang}, {and} \bibinfo{person}{Yuexian Zou}.}
  \bibinfo{year}{2022}\natexlab{}.
\newblock \showarticletitle{Unsupervised pre-training for temporal action
  localization tasks}. In \bibinfo{booktitle}{\emph{Proceedings of the IEEE/CVF
  Conference on Computer Vision and Pattern Recognition}}.
  \bibinfo{pages}{14031--14041}.
\newblock


\bibitem[Zhang et~al\mbox{.}(2018)]%
        {zhang2018grounding}
\bibfield{author}{\bibinfo{person}{Hanwang Zhang}, \bibinfo{person}{Yulei Niu},
  {and} \bibinfo{person}{Shih-Fu Chang}.} \bibinfo{year}{2018}\natexlab{}.
\newblock \showarticletitle{Grounding referring expressions in images by
  variational context}. In \bibinfo{booktitle}{\emph{Proceedings of the IEEE
  Conference on Computer Vision and Pattern Recognition}}.
  \bibinfo{pages}{4158--4166}.
\newblock


\bibitem[Zhang et~al\mbox{.}(2020a)]%
        {zhang2020learning}
\bibfield{author}{\bibinfo{person}{Songyang Zhang}, \bibinfo{person}{Houwen
  Peng}, \bibinfo{person}{Jianlong Fu}, {and} \bibinfo{person}{Jiebo Luo}.}
  \bibinfo{year}{2020}\natexlab{a}.
\newblock \showarticletitle{Learning 2d temporal adjacent networks for moment
  localization with natural language}. In \bibinfo{booktitle}{\emph{Proceedings
  of the AAAI Conference on Artificial Intelligence}},
  Vol.~\bibinfo{volume}{34}. \bibinfo{pages}{12870--12877}.
\newblock


\bibitem[Zhang et~al\mbox{.}(2020b)]%
        {zhang2020does}
\bibfield{author}{\bibinfo{person}{Zhu Zhang}, \bibinfo{person}{Zhou Zhao},
  \bibinfo{person}{Yang Zhao}, \bibinfo{person}{Qi Wang},
  \bibinfo{person}{Huasheng Liu}, {and} \bibinfo{person}{Lianli Gao}.}
  \bibinfo{year}{2020}\natexlab{b}.
\newblock \showarticletitle{Where does it exist: Spatio-temporal video
  grounding for multi-form sentences}. In \bibinfo{booktitle}{\emph{Proceedings
  of the IEEE/CVF Conference on Computer Vision and Pattern Recognition}}.
  \bibinfo{pages}{10668--10677}.
\newblock


\bibitem[Zhou et~al\mbox{.}(2018)]%
        {zhou2018weakly}
\bibfield{author}{\bibinfo{person}{Luowei Zhou}, \bibinfo{person}{Nathan
  Louis}, {and} \bibinfo{person}{Jason~J Corso}.}
  \bibinfo{year}{2018}\natexlab{}.
\newblock \showarticletitle{Weakly-supervised video object grounding from text
  by loss weighting and object interaction}.
\newblock \bibinfo{journal}{\emph{arXiv preprint arXiv:1805.02834}}
  (\bibinfo{year}{2018}).
\newblock


\bibitem[Zhu et~al\mbox{.}(2022)]%
        {zhu2022seqtr}
\bibfield{author}{\bibinfo{person}{Chaoyang Zhu}, \bibinfo{person}{Yiyi Zhou},
  \bibinfo{person}{Yunhang Shen}, \bibinfo{person}{Gen Luo},
  \bibinfo{person}{Xingjia Pan}, \bibinfo{person}{Mingbao Lin},
  \bibinfo{person}{Chao Chen}, \bibinfo{person}{Liujuan Cao},
  \bibinfo{person}{Xiaoshuai Sun}, {and} \bibinfo{person}{Rongrong Ji}.}
  \bibinfo{year}{2022}\natexlab{}.
\newblock \showarticletitle{Seqtr: A simple yet universal network for visual
  grounding}. In \bibinfo{booktitle}{\emph{Computer Vision--ECCV 2022: 17th
  European Conference, Tel Aviv, Israel, October 23--27, 2022, Proceedings,
  Part XXXV}}. Springer, \bibinfo{pages}{598--615}.
\newblock


\bibitem[Zhuang et~al\mbox{.}(2018)]%
        {zhuang2018parallel}
\bibfield{author}{\bibinfo{person}{Bohan Zhuang}, \bibinfo{person}{Qi Wu},
  \bibinfo{person}{Chunhua Shen}, \bibinfo{person}{Ian Reid}, {and}
  \bibinfo{person}{Anton Van Den~Hengel}.} \bibinfo{year}{2018}\natexlab{}.
\newblock \showarticletitle{Parallel attention: A unified framework for visual
  object discovery through dialogs and queries}. In
  \bibinfo{booktitle}{\emph{Proceedings of the IEEE Conference on Computer
  Vision and Pattern Recognition}}. \bibinfo{pages}{4252--4261}.
\newblock


\end{thebibliography}

\end{document}